\documentclass{article}

\usepackage{arxiv}

\usepackage[utf8]{inputenc}
\usepackage[T1]{fontenc}
\DeclareUnicodeCharacter{2011}{-}

\usepackage{multirow}
\usepackage{tablefootnote}
\usepackage{tabularx}
\usepackage{booktabs,colortbl}
\usepackage{longtable}
\usepackage{float}
\usepackage{pifont}
\usepackage{refcount}
\usepackage{subcaption}
\usepackage{xcolor}
\usepackage{xspace}
\usepackage{xparse}
\usepackage{etoolbox}
\usepackage[numbers]{natbib}
\usepackage{url}
\usepackage{amsfonts}
\usepackage{amsmath}
\usepackage{microtype}
\usepackage{graphicx}
\usepackage{hyperref}
\usepackage[capitalize,noabbrev]{cleveref}

\graphicspath{{./}}

\crefname{appendix}{Appendix}{Appendices}
\Crefname{appendix}{Appendix}{Appendices}

\providecommand{\Description}[1]{}

\newcommand{\shimmer}[1]{\cellcolor{teal!15}\textbf{#1}}

\newcommand{\cmark}{\ding{51}}
\newcommand{\xmark}{\ding{55}}

\newcommand{\tfn}[2]{\tablefootnote{\label{#1}#2}}

\newcommand{\tfnmark}[1]{\footnotemark[\getrefnumber{#1}]}

\newcommand{\numdatasets}{29\xspace}
\newcommand{\numpretraindatasets}{20\xspace}
\newcommand{\numdownstreamdatasets}{9\xspace}
\newcommand{\dataname}{EDAMAME\xspace}
\newcommand{\datanamefull}{\textbf{EDA} \textbf{M}ulti-dataset \textbf{A}rchive for \textbf{M}odel training and \textbf{E}valuation\xspace}
\newcommand{\modelname}{UME\xspace}
\newcommand{\modelnamefull}{fo\textbf{U}ndation \textbf{M}odel for \textbf{E}lectrodermal activity data\xspace}

\AtBeginDocument{%
  }


\begin{document}

\title{A Foundation Model for Electrodermal Activity Data}

\author{
Leonardo Alchieri\thanks{Lead and corresponding author.} \\
  Università della Svizzera italiana (USI) \\
  Via la Santa 1 \\
  6962 Lugano, Switzerland \\
  \texttt{leonardo.alchieri@usi.ch} \\
  \And
Matteo Garzon\thanks{Contributed to the preparation and curation of the \dataname dataset.} \\
  Università della Svizzera italiana (USI) \\
  Via la Santa 1 \\
  6962 Lugano, Switzerland \\
  \texttt{matteo.garzon@usi.ch} \\
  \And
Lidia Alecci\thanks{Contributed to the technical discussions and writing process.} \\
  Università della Svizzera italiana (USI) \\
  Via la Santa 1 \\
  6962 Lugano, Switzerland \\
  \texttt{lidia.alecci@usi.ch} \\
  \And
Francesco Bombassei De Bona\footnotemark[3] \\
  Università della Svizzera italiana (USI) \\
  Via la Santa 1 \\
  6962 Lugano, Switzerland \\
  \texttt{francesco.bombassei.de.bona@usi.ch} \\
  \And
Martin Gjoreski\footnotemark[3] \\
  Università della Svizzera italiana (USI) \\
  Via la Santa 1 \\
  6962 Lugano, Switzerland \\
  \texttt{martin.gjoreski@usi.ch} \\
  \And
Giovanni De Felice\thanks{Supervised the scientific and writing aspects.} \\
  Università della Svizzera italiana (USI) \\
  Via la Santa 1 \\
  6962 Lugano, Switzerland \\
  \texttt{giovanni.de.felice@usi.ch} \\
  \And
Silvia Santini\footnotemark[4] \\
  Università della Svizzera italiana (USI) \\
  Via la Santa 1 \\
  6962 Lugano, Switzerland \\
  \texttt{silvia.santini@usi.ch} \\
}

\maketitle

\begin{abstract}
Foundation models have recently extended beyond natural language and vision to time‑series domains, including physiological signals. Yet developing foundation models for electrodermal activity (EDA) remains hindered by the absence of large‑scale, curated, openly accessible datasets. EDA reflects sympathetic activity and is widely used to infer cognitive load, stress, and engagement. However, few wearable devices feature EDA sensing, and the only large‑scale EDA archive to date is proprietary. To address this gap, we compile \dataname, a collection of EDA traces from \numdatasets public datasets comprising approximately 23{,}000 hours of data from 731 users. Using this collection, we train \modelname, the first foundation model for EDA.
As a frozen feature extractor, \modelname outperforms generic handcrafted features and matches generalist time-series foundation models while requiring $20\times$ fewer computational resources; fine-tuning on the downstream tasks further yields additional performance gains in most experiments. \modelname also generalizes to EDA from two devices unseen during training.
 We further show that \modelname's performance improves as the amount of pretraining data grows, underscoring the importance of large-scale data for effective foundation models. Our findings highlight intrinsic challenges of EDA signals, revealing future research opportunities.
We share \dataname under tiered access and also release model weights and code.
\end{abstract}


\section{Introduction}
\label{sec:introduction}
    Thanks to their ability to learn general patterns from broad, diverse data and to adapt to a wide range of downstream tasks, foundation models have attracted attention beyond their traditional applications in natural language processing (NLP) and computer vision. Generalist foundation models for \emph{time series}, such as Mantis~\cite{Feofanov2025} and Chronos~\cite{Ansari2024,Ansari2025}, have been shown to achieve remarkable performance on a diverse range of downstream tasks. Emerging foundation models for \emph{physiological time series} data -- e.g., for photoplethysmogram (PPG)~\cite{Abbaspourazad2024,Pillai2025,Saha2025,Ding2024,Luo2025}, electrocardiogram (ECG)~\cite{Abbaspourazad2024,McKeen2025,Li2025}, and electroencephalogram (EEG) data~\cite{Sukhbaatar2025} -- show compelling performance, too. Yet their training remains constrained by the absence of large‑scale, curated, and open datasets for physiological signals. As highlighted by \citet{Abbaspourazad2024} with respect to time series data in the medical domain, ``\emph{datasets are usually small in comparison to other domains, which is an obstacle for developing neural network models for biosignals}''.

    This paucity of data is particularly acute for electrodermal activity (EDA) signals. EDA refers to changes in the skin’s electrical conductance caused by variations in sweat gland activity, which is itself regulated by the sympathetic branch of the autonomic nervous system (ANS).
    Because EDA increases with physiological arousal, it is commonly used to assess cognitive load~\cite{kahneman1969pupillary,romine2020using}, stress~\cite{gashiDetectionArtifactsAmbulatory2020,pinge2024detection}, and engagement~\cite{Gao2020,Gashi2019,bustos2022wearables}. However, EDA sensors are not (yet) commonplace on wearable devices and only a few (e.g., the Empatica E4 wristband) feature the unobtrusive, continuous measurement modality necessary to collect longitudinal traces in real-world settings. The only large-scale archive of EDA data is a proprietary dataset collected using the Fitbit Sense 2~\cite{McDuff2024.02.22.581472}. This lack of open data has hampered the development of foundation models for EDA.

    To cope with this challenge, we first constructed \dataname (\datanamefull), a large-scale archive of EDA data that integrates records from \numdatasets publicly available datasets, totalling almost 23,000 hours of data from 731 users.
    Using \dataname, we were able to train \modelname (\modelnamefull),  to the best of our knowledge the first foundation model for EDA.

    We evaluated \modelname on several downstream tasks and found that it outperforms baseline models trained on generic handcrafted features and performs comparably to models trained on EDA-specific handcrafted features. It also matches the performance of larger, generalist time-series foundation models while requiring at least $20\times$ fewer computational resources (in GFLOPs).
     We also show that \modelname's representations generalize to two wearable EDA sensors unseen during pretraining, and that fine-tuning further improves performance on the downstream tasks.
    Scaling‑law analyses, following ~\citet{Narayanswamy2024}, show that \modelname's performance improves as the amount of pretraining data grows, indicating that a dataset of the scale of \dataname is needed to achieve these results.
    Our results, however, also highlight the intrinsic difficulty of working with EDA signals: balanced accuracy scores rarely exceed 0.7 and exhibit substantial variability across tasks and users. More research is therefore needed to fully harness the potential of EDA for both unimodal and multimodal ubiquitous sensing.

    To support the research community, we release as open source the full Python codebase developed for this work, along with \modelname's code and pretrained weights.
    We also release the \dataname dataset collection under a tiered scheme: an openly accessible version comprising 16 open-license datasets, and a complete version that adds 11 further datasets and requires signing a data sharing agreement. The remaining two datasets are available only through the original distribution channels; for these we release the code needed to reproduce our preprocessing.
    Details on data and code availability are provided in \cref{sec:dataset_code_avail}.

\section{Background and Related Work}
    In this section, we first provide background on wearable devices and EDA data.
    We then review existing work on foundation models for physiological data, focusing on signals collected from wearable devices.
    Finally, we discuss self-supervised learning applied to EDA data, since it is one of the building blocks for training foundation models.

    \paragraph{Background}
    Foundation models require large-scale datasets to be trained~\cite{Bommasani2022}. In NLP, state-of-the-art results~ \cite{Apertus2025,Brown2020} have been enabled in part by the availability of massive text corpora. Such corpora can be built, for instance, by crawling the Web to gather publicly available text, as done by the Common Crawl project\footnote{\url{https://commoncrawl.org}}. Collecting comparably large collections of physiological signals is far more challenging, as it requires specialized equipment and substantial  human effort and time~\cite{Liu2025,Slade2025}.
    As a result, foundation models for physiological data have so far relied largely on clinical corpora~\cite{Lee2025}. Only recent work, e.g., \cite{Abbaspourazad2024}, explores the use of foundation models trained on real-world wearable physiological signals.
    This is because high-quality wearable data is significantly harder to collect compared to clinical data~\cite{Bizzego2020}.
    Moreover, despite its use in real settings such as seizure detection~\cite{CasanovasOrtega2022}, EDA is collected in clinical settings far less often than PPG or ECG~\cite{Coffman2020,Poh2012}, so clinical corpora offer limited coverage of this modality.

    \paragraph{Foundation models for physiological data}
    \cref{tab:foundation_paper_overview} lists a selection of existing foundation models for physiological data, along with information about their size, availability, and pretraining data. This list, albeit not exhaustive, shows that current research focuses on PPG, ECG, or multimodal approaches, and that most models rely on clinical data.

    \citet{Abbaspourazad2024} were the first to train foundation models for PPG and ECG on a large-scale private dataset collected with wearable devices. They showed that the learned representations could be used for multiple downstream tasks, with performance on par with existing approaches.
    Following these results, other foundation models for either PPG or ECG data were developed, e.g., \cite{Saha2025,Pillai2025,Ding2024,McKeen2025,Li2025}. However, none of them targeted EDA and, to the best of our knowledge, \citet{Saha2025} is the only work that trains an open-source foundation model exclusively on wearable physiological data.

    Multimodal approaches have also recently been proposed.
    \citet{Narayanswamy2024} trained a family of multimodal foundation models on aggregated features computed at 1-minute granularity for PPG (e.g., mean heart rate), EDA (e.g., mean EDA values), and other physiological and behavioral signals. These models were found to outperform baseline methods by 50\% (area under the receiver operating characteristic curve). However, they operate on aggregated features rather than raw signals, rely on proprietary data, and neither the code nor the model weights have been released.
    Using publicly available data, \citet{Luo2025} trained a multimodal foundation model for physiological signals on a curated collection of about 15,000 hours of traces, including PPG, ECG, and EDA, and reported promising zero-shot downstream performance.
    However, EDA accounts for less than 10\% (approximately 1,500 hours) of the total amount of data in their collection.
    Overall, a curated EDA dataset collection has not yet been made available to the wider research community.

    \begin{table}[t]
        \centering
        \caption{Overview of selected foundation models for physiological data along with information about the data used to train them.}
        \resizebox{\textwidth}{!}{%
        \begin{tabular}{ccccc|ccccc}
        \toprule
            \multicolumn{5}{c|}{\textbf{Foundation model}} & \multicolumn{5}{c}{\textbf{Train data}} \\
             \textbf{Ref.} & \textbf{Year} & \textbf{Signal(s)} & \textbf{Params.} & \textbf{Avail.} & \textbf{Dataset(s)} & \textbf{Data type} & \textbf{Hours (k)} & \textbf{Users} & \textbf{Avail.}  \\
             \midrule
             \addlinespace[6pt]
                \rowcolor{gray!12}
                \multicolumn{10}{l}{\textbf{\scshape Clinical data}} \\
                \addlinespace[3pt]

             \multirow{3}{*}{\cite{Pillai2025}} & \multirow{3}{*}{2025} & \multirow{3}{*}{PPG} &  \multirow{3}{*}{$\sim$ 30 - 150 M} & \multirow{3}{*}{Open source}
                   &  VitalDB~\cite{Lee2022} & Clinical & $\sim$ 17 & $\sim$ 6k  & \cmark \\
             & & & & &  MIMIC-III~\cite{Johnson2016} & Clinical & $\sim$ 20 & $\sim$ 6k  & \cmark \\
             & & & & &  MESA~\cite{Zhang2018} & Clinical & $\sim$ 20 & $\sim$ 2k  & \cmark \\

             \cite{Ding2024} & 2024 & PPG & $\sim$ 1-8 M & Code only & \cite{Ding2024} & Clinical & $\sim$ 300 & $\sim$ 29k  & \cmark \\

             \multirow{3}{*}{\cite{McKeen2025}} & \multirow{3}{*}{2025} & \multirow{3}{*}{ECG} & \multirow{3}{*}{$\sim$ 300 M} & \multirow{3}{*}{Open source} & MIMIC IV~\cite{Johnson2023} & Clinical & $\sim$ 140 & $\sim$ 160k  & \cmark \\
             & & & & & PhysioNet2021~\cite{Reyna2021} & Clinical & $\sim$ 14 & N/A & \cmark \\
             & & & & & UHN-ECG~\cite{McKeen2025} & Clinical & $\sim$ 100 & $\sim$ 180k   & \cmark \\

             \cite{Li2025} & 2025 & ECG & $\sim$ 30M  & Open source & HEEDB~\cite{Koscova} & Clinical & $\sim$ 1,000 & $\sim$ 2M  & \cmark \\

            \cite{Sukhbaatar2025} & 2025 & EEG & not stated & Open source & curated collection & Clinical & $\sim$ 350 & $\sim$ 9  & \xmark\tfn{fn:collection}{Individual datasets available, but not the curated collection.} \\

            \addlinespace[6pt]
                \rowcolor{gray!12}
                \multicolumn{10}{l}{\textbf{\scshape Wearable data}} \\
                \addlinespace[3pt]
             \multirow{2}{*}{\cite{Abbaspourazad2024}} & \multirow{2}{*}{2024} & PPG & $\sim$ 3M & Private & \multirow{2}{*}{AHMS~\cite{Truslow2024}} & \multirow{2}{*}{Wearable} & $\sim$ 300 & $\sim$ 141k  & \xmark \\
             & & ECG & $\sim$ 3M & Private & & & $\sim$ 30 & $\sim$ 106k  & \xmark \\

            \cite{Narayanswamy2024} & 2024 & multimodal & $\sim$ 1-100M & Private & \cite{Narayanswamy2024} & Wearable & $\sim$ 40,000 & $\sim$ 165k  & \xmark \\

             \cite{Saha2025} & 2025 & PPG & $\sim$ 30M & Open source & MOODS~\cite{Neupane2024} & Wearable & $\sim$ 40 & $\sim$ 120  & \xmark \\

             \addlinespace[6pt]
                \rowcolor{gray!12}
                \multicolumn{10}{l}{\textbf{\scshape Other}} \\
                \addlinespace[3pt]

            \cite{Luo2025} & 2025 & multimodal & $\sim$ 140M & Open source & curated collection & Mixed & $\sim$ 15 & N/A & \xmark\tfnmark{fn:collection} \\

             \addlinespace[6pt]
                \rowcolor{gray!12}
                \multicolumn{10}{l}{\textbf{\scshape Ours}} \\
                \addlinespace[3pt]
             \textbf{Ours} & \textbf{2026} & \textbf{EDA} & $\sim$ \textbf{1M} & \textbf{Open source} & \textbf{EDAMAME} & \textbf{Wearable} & $\sim$ \textbf{20} & $\sim$ \textbf{440} & \textbf{\cmark} \\
             \bottomrule
        \end{tabular}%
        }

        \label{tab:foundation_paper_overview}
    \end{table}

    \paragraph{Self-supervised learning for EDA data}
    Researchers have also explored the use of self-supervised learning on wearable data, including EDA. \citet{Dissanayake2022} trained a self-supervised learning model, through contrastive learning, for emotion estimation using PPG, EDA and skin temperature. \citet{Saeed2021} similarly used a multimodal self-supervised approach in a federated learning setting. Recently, \citet{Matton2023} trained a self-supervised learning model on EDA data to perform stress classification. They used contrastive learning and data augmentation techniques to achieve state-of-the-art performance on in-distribution downstream tasks.

    \vspace{0.5\baselineskip}
    Overall, existing work suggests that foundation models for physiological data can be trained and used on a diverse set of downstream tasks when using PPG, ECG or multimodal approaches. However, many publicly available physiological foundation models rely on clinical data, and there is limited evidence that features from these models can be used with data from wearable devices and real-life settings.
    Recent work on multimodal~\cite{Luo2025} and PPG~\cite{Saha2025} foundation models shows that, even without large-scale proprietary datasets, researchers are able to train open source foundation models. However, this work does not provide a readily available collection of datasets for researchers to build upon, hindering the development of wearable foundation models.
    Given promising work on self-supervised learning for EDA data~\cite{Matton2023}, we create a collection of datasets, \dataname, which we make available to the research community and, with it, we train an open source foundation model for EDA data, \modelname, whose code and weights we also release.

\section{\dataname: A Collection of Electrodermal Activity Datasets}
\label{sec:edamame}

    In this section, we describe how we address the lack of large-scale datasets containing wearable EDA data through \dataname (\datanamefull). \dataname is a collection of existing, smaller-scale, datasets prepared and pre-processed in a unified manner. The goal of \dataname is to enable researchers to train foundation models for wearable EDA data.

    \subsection{Description of the Collection of Datasets}

        Our goal is to create a large-scale and diverse corpus to address the scarcity of EDA-specific collections.
        To this end, we first identify a set of scenarios that are relevant for EDA data from the literature~\cite{Hossain2024}.
        The scenarios are: \emph{sleep monitoring}, \emph{stress/emotion induction}, \emph{engagement}, \emph{real-world stress}, \emph{workplace analysis}, and \emph{daily living}.
        Then, we search and select datasets from the literature using the following criteria:
        \begin{enumerate}
            \item datasets have to contain \emph{raw} EDA data from wearable devices;
            \item the individual datasets have to be either open source or available to researchers upon signing a data sharing agreement;
            \item datasets have to contain data collected during tasks or moments that influence EDA signals;
            \item the collection has to contain datasets collected using different protocols, e.g., lab environment or in-the-wild collection;
            \item the collection needs to contain at least one dataset from each one of the six scenarios defined;
            \item in order to limit the search size, we also add a saturation criterion: once we reach the threshold of 100 users \emph{or} 1000 hours for one of the six scenarios, we stop searching for additional data;
            \item the total size of \dataname has to be more than 15,000 hours and more than 100 users, which is a size similar to that of datasets used to train existing open source foundation models for wearable data~\cite{Saha2025,Luo2025}.
        \end{enumerate}

        We search for datasets through open source databases containing physiological data, specifically PhysioNet\footnote{\url{https://physionet.org/}}, Zenodo\footnote{\url{https://zenodo.org/}}, and Kaggle\footnote{\url{https://www.kaggle.com}}.
        We also search for dataset information using databases of scientific articles, e.g., Google Scholar\footnote{\url{https://scholar.google.com}}, ACM Digital Library\footnote{\url{https://dl.acm.org}}.
        Finally, we also search through the online databases of scientific journals that publish datasets containing physiological data, e.g., Nature Scientific Data\footnote{\url{https://www.nature.com/sdata/}}, IMWUT\footnote{\url{https://dl.acm.org/journal/imwut}}.
        We identify a potential 40 datasets. Through the aforementioned criteria, we select a total of \textbf{\numdatasets} datasets for \dataname.
        We report in \cref{tab:pretraining_datasets} an overview of the \numdatasets datasets.
        In total, \dataname contains approximately 23,000 hours of EDA data from 731 users.
        The size of \dataname is in line, as outlined by our seventh selection criterion, with collections of datasets used by \citet{Saha2025} and \citet{Luo2025} to train open source foundation models for wearable physiological data.
        Moreover, scaling‑law experiments (\cref{app:dataset_size}) confirm that the size of \dataname allows us to achieve performance comparable to or exceeding existing methods.

        Beyond these criteria, we adopt a \emph{device-stratified} design, which we summarize in \cref{tab:pretraining_datasets}. The pretraining split, \dataname-train, contains only datasets recorded with the Empatica E4 wristband\footnote{\url{https://www.empatica.com/research/e4/}}. We restrict pretraining to a single device for two reasons: the E4 is the research-grade wearable with continuous EDA sensing for which by far the largest amount of public data exists, and keeping the pretraining corpus device-pure removes hardware-specific domain shifts, so that the model is encouraged to learn physiological structure rather than sensor idiosyncrasies. The downstream evaluation split, \dataname-test, instead deliberately mixes devices: it contains data from the Empatica E4 as well as from two Shimmer units, the Shimmer 2R and the Shimmer3 GSR+. This lets us quantify whether \modelname transfers to EDA recorded with hardware never seen during pretraining, which we report in \cref{sec:comparison_complete}.
        We select these three non-Empatica datasets because they are publicly available and provide labels for which EDA is a plausible predictor. Specifically, we use the artifact annotations of EDABE~\cite{llanes2023edabe}, whose automatic detection the community has addressed with machine learning~\cite{gashiDetectionArtifactsAmbulatory2020}; the relaxation-versus-video contrast of Multi-physio~\cite{lucas2026physiological}, which we take as a proxy for cognitive load and emotional engagement; and the self-reported arousal and liking labels of Amigos~\cite{miranda2018amigos}.

        We select a common sampling rate of $4~\mathrm{Hz}$ for EDA data processed for \dataname, since EDA data from Empatica E4 devices is already sampled at $4~\mathrm{Hz}$\footnote{\url{https://www.empatica.com/blog/decoding-wearable-sensor-signals-what-to-expect-from-your-e4-data}}. We downsample data from other devices (Shimmer 2R and Shimmer3 GSR+) to $4~\mathrm{Hz}$.
        All timestamps are converted to UTC time. Whenever timestamps are missing, we assign Unix time 0 to the start of a time series of EDA data.

        \begin{table}[t]
            \centering
            \footnotesize
            \caption{Summary of all datasets included in the EDAMAME collection, including the amount of EDA data in terms of hours, the number of users, the scenario (among the six identified), whether the data were collected in the lab or in the wild, the device used to record the EDA data, and the split the dataset belongs to. \emph{Train} denotes membership in \dataname-train, used exclusively to pretrain \modelname; \emph{Test} denotes membership in \dataname-test, used exclusively for downstream evaluation. Note that every dataset in \dataname-train was recorded with the Empatica E4: no Shimmer data enters pretraining. Durations are computed from the number of EDA samples in the prepared files, at 4~Hz.}
            \label{tab:pretraining_datasets}
            \setlength{\tabcolsep}{3.5pt}

            \begin{tabular}{lrrlllc}
                \toprule
                \textbf{Dataset Name} & \textbf{Duration (h)} & \textbf{\# Users} & \textbf{Scenario} & \textbf{Environment} & \textbf{Device} & \textbf{Split} \\
                \midrule
                \rowcolor{red!5} Amigos~\cite{miranda2018amigos} & 22.18 & 15 & Stress/Emotion Induction & Lab & Shimmer 2R & Test \\
                \rowcolor{red!5} APSync~\cite{Gashi2019} & 167.75 & 27 & Real-world Stress & Wild & Empatica E4 & Test \\
                \rowcolor{blue!5} BIG IDEAS~\cite{Bent2021} & 2607.19 & 16 & Daily Living & Wild & Empatica E4 & Train \\
                \rowcolor{blue!5} BiHeartS~\cite{Abdalazim2025} & 2912.57 & 11 & Sleep Monitoring & Wild & Empatica E4 & Train \\
                \rowcolor{red!5} DREAMT~\cite{Wang} & 883.39 & 100 & Sleep monitoring & Wild & Empatica E4 & Test \\
                \rowcolor{blue!5} Dynamics in the workplace~\cite{Lukan2018} & 1813.86 & 42 & Workplace Analysis & Wild & Empatica E4 & Train \\
                \rowcolor{red!5} EDABE~\cite{llanes2023edabe} & 74.46 & 43 & Daily Living & Wild & Shimmer3 GSR+ & Test \\
                \rowcolor{blue!5} EmoStress~\cite{gjoreski2017monitoring,gjoreski2016continuous} & 1730.60 & 5 & Real-world Stress & Wild & Empatica E4 & Train \\
                \rowcolor{blue!5} EmpaticaE4Stress~\cite{Campanella2023} & 18.02 & 29 & Stress/Emotion Induction & Lab & Empatica E4 & Train \\
                \rowcolor{blue!5} EPM-E4~\cite{Garcia-Moreno2020} & 21.51 & 47 & Engagement & Wild & Empatica E4 & Train \\
                \rowcolor{red!5} HeartS~\cite{Abdalazim2023} & 887.07 & 5 & Sleep Monitoring & Wild & Empatica E4 & Test \\
                \rowcolor{red!5} HHISS~\cite{xiao2025hhiss} & 22.13 & 46 & Real-world Stress & Wild & Empatica E4 & Test \\
                \rowcolor{blue!5} LAUREATE~\cite{Laporte2023} & 1406.71 & 46 & Engagement & Wild & Empatica E4 & Train \\
                \rowcolor{blue!5} M2Sleep~\cite{Gashi2022a} & 8405.53 & 16 & Sleep Monitoring & Wild & Empatica E4 & Train \\
                \rowcolor{blue!5} MEFAR~\cite{Derdiyok2024} & 26.34 & 23 & Stress/Emotion Induction & Lab & Empatica E4 & Train \\
                \rowcolor{red!5} Multi-physio~\cite{lucas2026physiological} & 1.24 & 10 & Stress/Emotion Induction & Wild & Shimmer3 GSR+ & Test \\
                \rowcolor{blue!5} Nurses' Stress~\cite{Hosseini2022} & 99.91 & 18 & Workplace Analysis & Wild & Empatica E4 & Train \\
                \rowcolor{blue!5} PPG-Dalia~\cite{Reiss2019} & 37.70 & 15 & Daily Living & Wild & Empatica E4 & Train \\
                \rowcolor{blue!5} Regulation~\cite{azgomi2023regulation} & 60.87 & 20 & Stress/Emotion Induction & Lab & Empatica E4 & Train \\
                \rowcolor{blue!5} SEED~\cite{DiLascio2018} & 336.73 & 31 & Engagement & Wild & Empatica E4 & Train \\
                \rowcolor{blue!5} SEED-II-Lab~\cite{DiLascio2018} & 28.40 & 25 & Engagement & Lab & Empatica E4 & Train \\
                \rowcolor{blue!5} SEED-II-Wild~\cite{DiLascio2018} & 156.66 & 6 & Engagement & Wild & Empatica E4 & Train \\
                \rowcolor{blue!5} Stress Predict~\cite{Iqbal2022} & 32.24 & 35 & Stress/Emotion Induction & Lab & Empatica E4 & Train \\
                \rowcolor{blue!5} ToadStool~\cite{Svoren2020} & 9.45 & 10 & Stress/Emotion Induction & Lab & Empatica E4 & Train \\
                \rowcolor{red!5} USILaughs~\cite{DiLascio2019} & 29.29 & 34 & Stress/Emotion Induction & Lab & Empatica E4 & Test \\
                \rowcolor{blue!5} WEEE~\cite{Gashi2022} & 16.45 & 17 & Daily Living & Lab & Empatica E4 & Train \\
                \rowcolor{red!5} WESAD~\cite{Schmidt2018} & 29.71 & 15 & Stress/Emotion Induction & Lab & Empatica E4 & Test \\
                \rowcolor{blue!5} WESD~\cite{RafiulAmin2022} & 123.13 & 10 & Real-world Stress & Wild & Empatica E4 & Train \\
                \rowcolor{blue!5} Workplace~\cite{DiLascio2021} & 830.01 & 14 & Workplace Analysis & Wild & Empatica E4 & Train \\
                \midrule
                \rowcolor{blue!5} \emph{\dataname-train} (20) & 20,674 & 436 & & & & \\
                \rowcolor{red!5} \emph{\dataname-test} (9) & 2,117 & 295 & & & & \\
                \textbf{Total} & \textbf{22,791} & \textbf{731} & & & & \\
                \bottomrule
            \end{tabular}
        \end{table}

    \subsection{Data Variability in \dataname}
    \label{sec:data_variability}

\begin{figure}[t]
          \centering
          \includegraphics[width=0.7\linewidth]{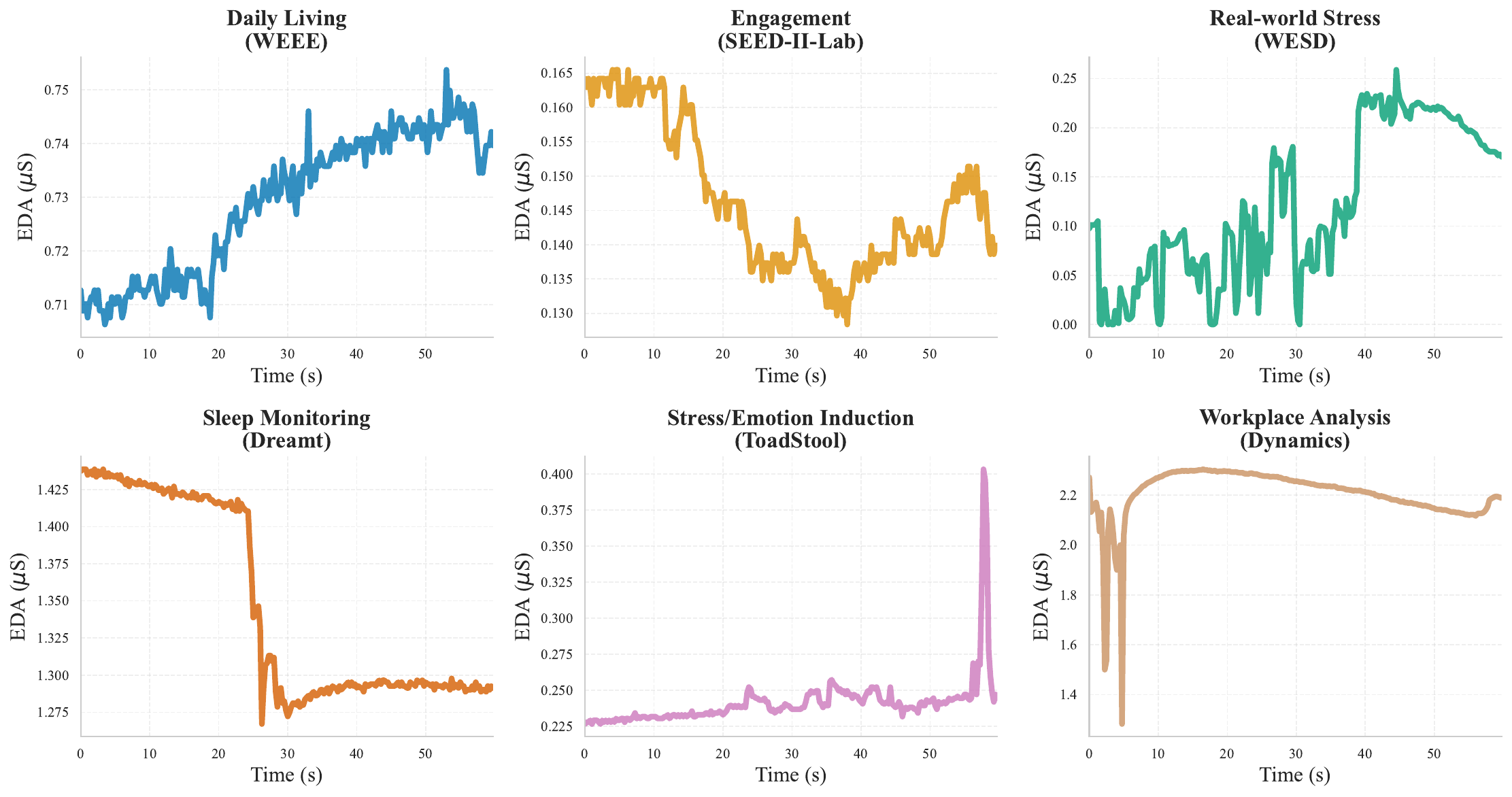}
          \caption{Example of six EDA signals (randomly drawn). Each signal is from a different dataset, each for one of the six scenarios.}
          \Description{Example of six EDA signals (randomly drawn).}
          \label{fig:examples}
        \end{figure}

        In this section, we characterize the variability of the EDA data in \dataname, which is what makes the collection suitable for training foundation models.
        In particular, we discuss the EDA data distribution and the distribution of users and data per user across the datasets.
        \cref{fig:examples} shows six example EDA signals from different datasets, representing six distinct scenarios: \emph{sleep monitoring}, \emph{stress/emotion induction}, \emph{engagement}, \emph{real-world stress}, \emph{workplace analysis}, and \emph{daily living}.

\begin{figure}[t]
          \centering
          \includegraphics[width=\linewidth]{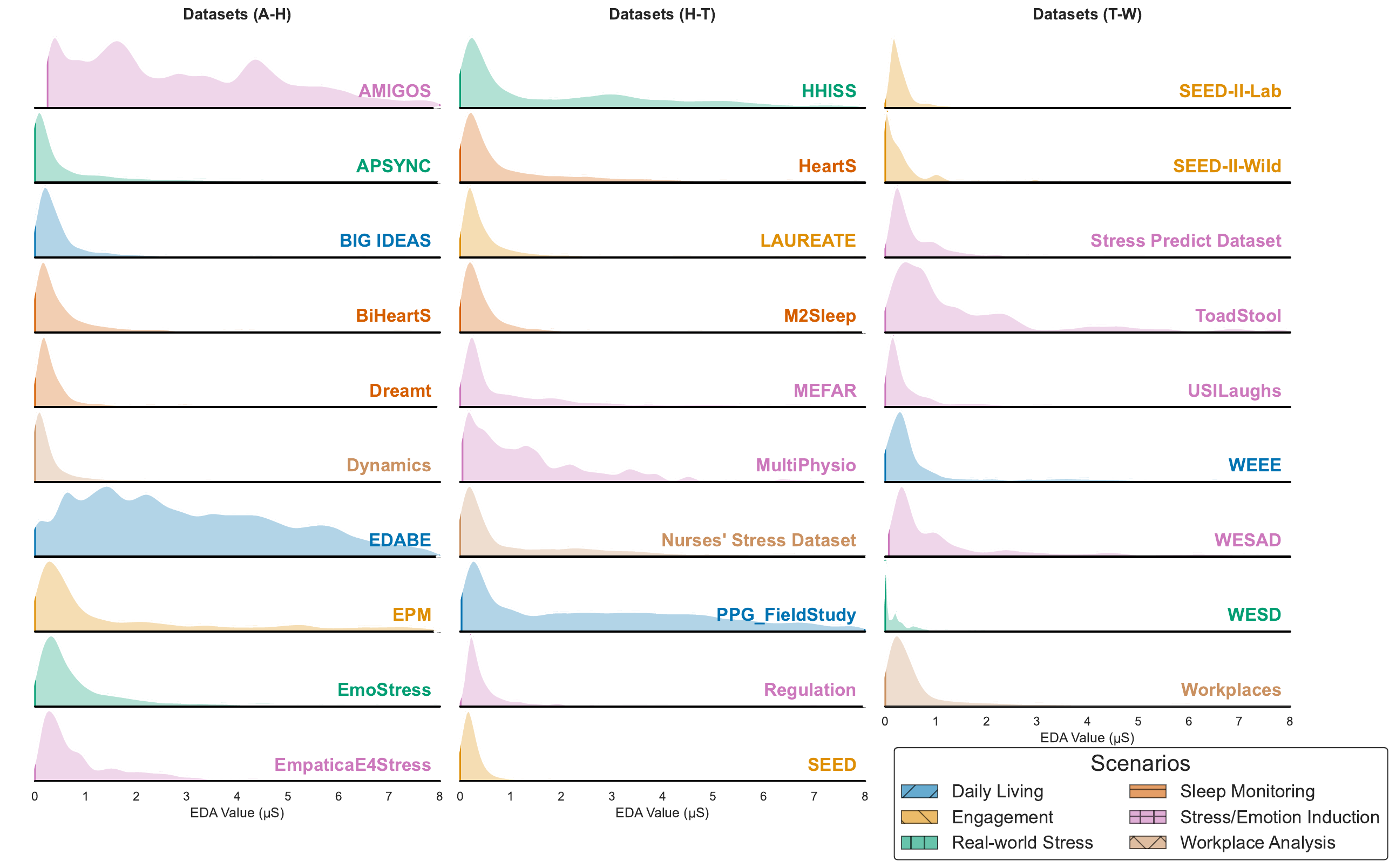}
          \caption{Ridgeline plot with distribution of EDA values from each dataset. The colors represent the six scenarios each dataset was collected in. The plot has been truncated, for the sake of visualization, to the range $0-8~\mu\mathrm{S}$. The original range was $0-40~\mu\mathrm{S}$ (the E4's max theoretical value is $100~\mu\mathrm{S}$).}
          \Description{Ridgeline plot with distribution of EDA values from each dataset.}
          \label{fig:rigdeline}
        \end{figure}

        \paragraph{Data distribution}
        \cref{fig:rigdeline} shows the distribution of raw EDA values across the \numdatasets datasets in \dataname.
        We highlight how data distribution is similar across datasets that are collected in similar scenarios.
        We note that higher EDA values are associated with stronger ANS arousal activation~\cite{Boucsein2012}.
        For example, datasets in the \emph{Stress/Emotion Induction} scenario have longer tails, in their distribution, compared to others.
        On the other hand, datasets in the \emph{Real-world Stress} scenario have lower tails and more EDA values close to $0~\mu\mathrm{S}$.
        These two distinct patterns highlight the challenges of real-world EDA data: a lower distribution of EDA values in the \emph{Real-world Stress} scenario suggests that during daily life fewer arousal-associated moments are present.
        Overall, \dataname contains diverse data, across both scenarios and EDA values.

\begin{figure}[t]
          \centering
          \includegraphics[width=0.5\linewidth]{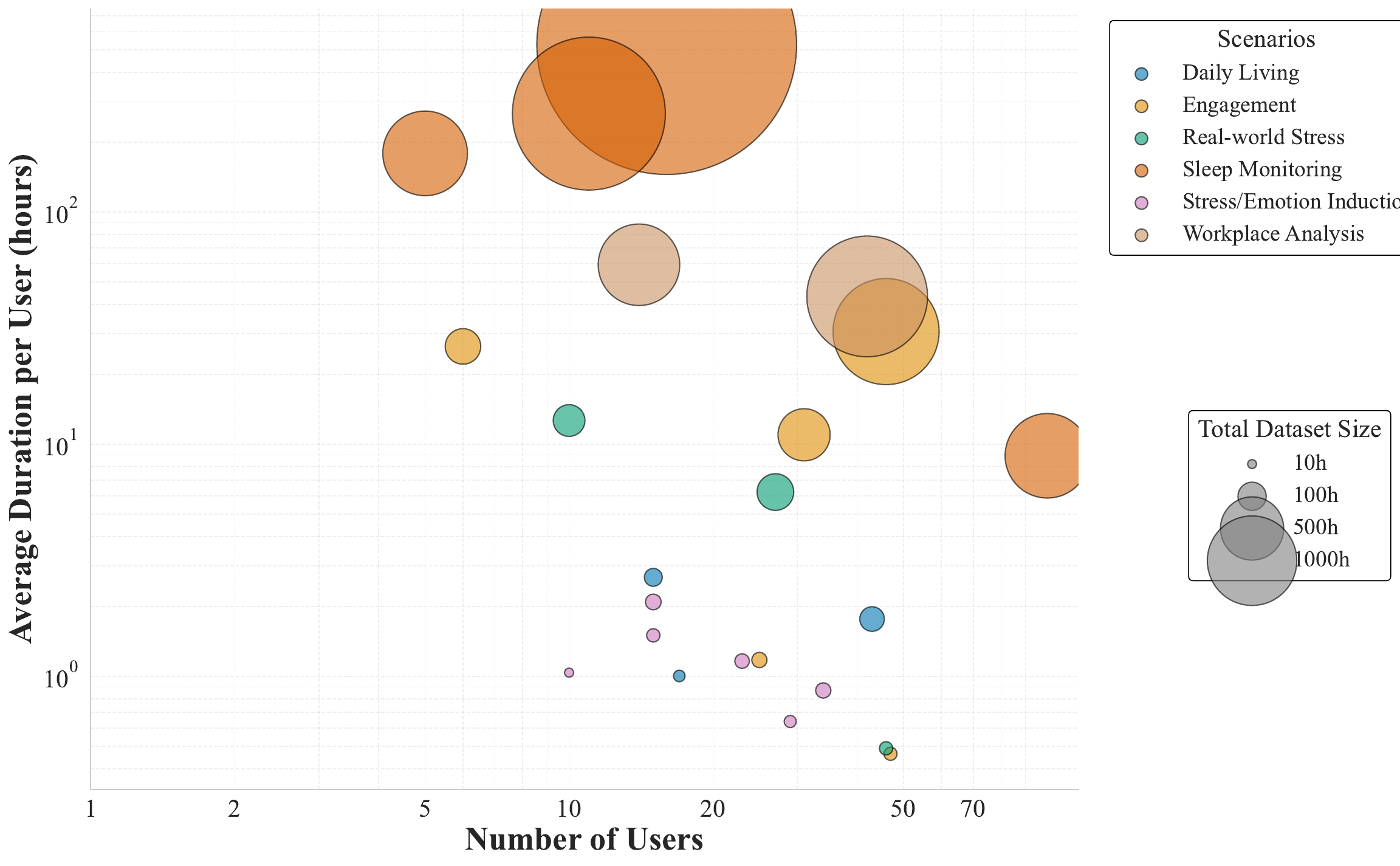}
          \caption{Number of users and data per user across the \dataname collection of datasets. Marker size is proportional to the size of the dataset (in terms of hours of EDA data).}
          \Description{Number of users and data per user across the \dataname collection of datasets.}
          \label{fig:num_users}
        \end{figure}

        \paragraph{Number of users and data per user}
        \cref{fig:num_users} shows the number of users compared to the data density per user, for each dataset in \dataname. The plot shows the structural topology of the collection of datasets. Specifically, the collection contains two types of datasets: datasets that contain fewer users but more data per user (longitudinal depth); and datasets that contain more users but less data per user (high population diversity). The first set of datasets in which multiple days of data for each user are present contains more intra-personal variability information, i.e., how people's data changes over time, compared to the second set. The second set, on the other hand, contains more inter-personal variability information, i.e., how data differs between different participants, compared to the first one.

        \vspace{0.5\baselineskip}
        \dataname contains data from a diverse range of scenarios and users. The collection also balances datasets with longitudinal depth and datasets with high population diversity. This diversity and variability are in line with recent literature on foundation model training data~\cite{Chen2025,Bukharin2024}.

\section{\modelname: An Open Source Foundation Model for EDA Data}

    \begin{figure}
        \centering
        \includegraphics[width=\linewidth]{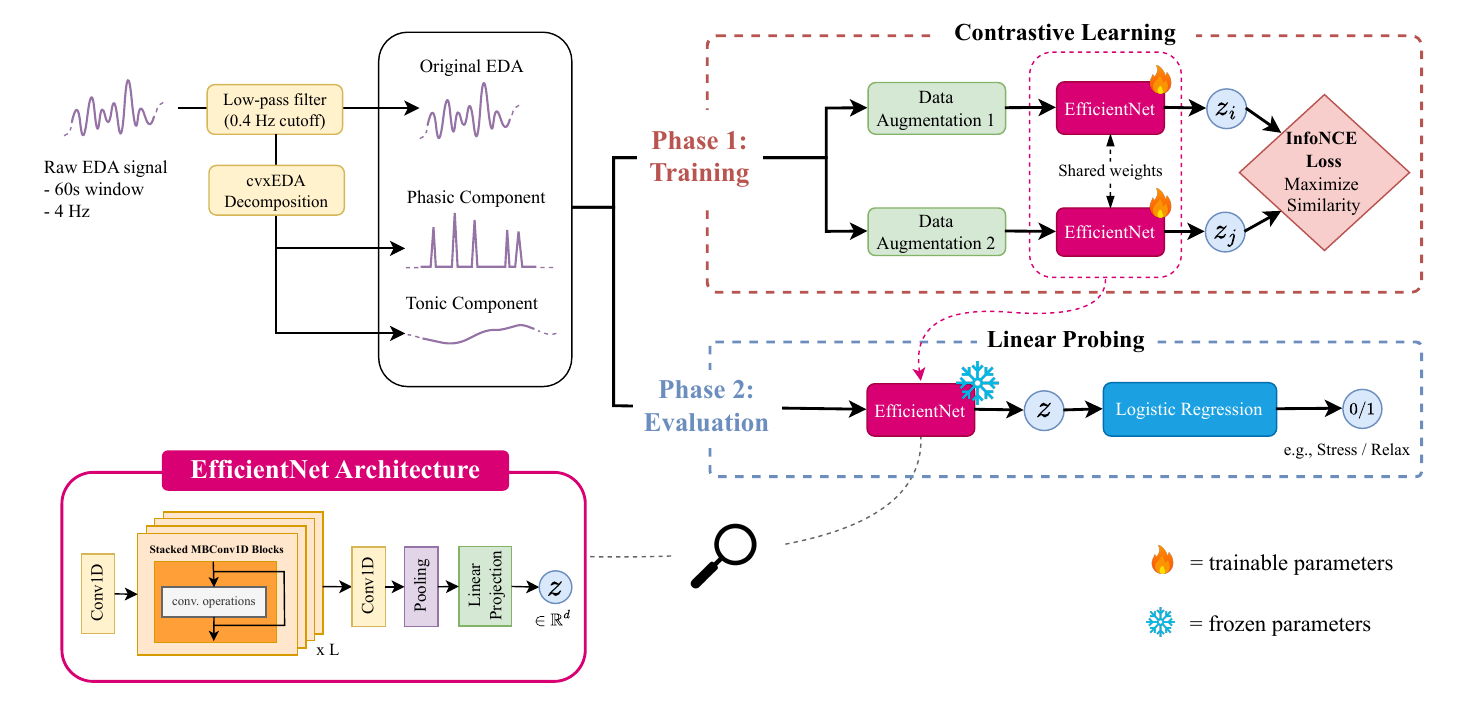}
        \caption{Overview of the training and downstream evaluation process we used for the \modelname foundation model. For reference, we report a sketch of the EfficientNet architecture, which we use as the backbone for \modelname. The dashed arrow indicates that the encoder pretrained in Phase 1 is the one used in Phase 2: for linear probing, shown here, its weights are frozen (snowflake) and only the logistic regression is fit; when instead fine-tuning (\cref{app:finetuning}), the same encoder is initialized from the Phase-1 weights and its parameters are updated together with the classification head. Not shown in this diagram is the second experiment we performed, which involves \emph{fine-tuning} the foundation model on the downstream tasks.}
        \Description{Overview of the training and downstream evaluation process we used for the \modelname foundation model.}
        \label{fig:overview}
    \end{figure}

    In this section, we describe \modelname (\modelnamefull), the first open source foundation model trained on wearable EDA data.
    In particular, we explain the data pre-processing, the training criteria, and the model architecture, and we provide additional information about training the model.
    With \modelname, our objective is to obtain a model whose internal representations encode key characteristics of the input EDA signal, such that these representations can be extracted and used as substitutes for domain-informed features in downstream tasks.
    To achieve this objective, as common for foundation models for physiological data, e.g., \cite{Abbaspourazad2024,Pillai2025,Saha2025}, we rely on self-supervised learning with a contrastive learning objective.
    To train \modelname we use a subset of the \dataname collection of datasets, consisting of around 20,000 hours of EDA data from approximately 440 users.
    We release as open source the full codebase and pretrained weights of the model, and make the \dataname dataset collection available to the research community (see \cref{sec:dataset_code_avail} for details).
    We show in \cref{fig:overview} an overview of the pipeline used to train and evaluate \modelname.

    \subsection{Data Pre-Processing and Preparation}

        \paragraph{Data split}
        We divide \dataname into two parts: a train part and a downstream evaluation part.
        For simplicity, we call the first \dataname-train and the second \dataname-test.
        We use \dataname-train to train our foundation model for wearable EDA data, and \dataname-test to perform evaluation on a set of downstream classification tasks.
        We select \numpretraindatasets datasets for \dataname-train and \numdownstreamdatasets datasets for \dataname-test.
        We perform this split to allow models to be tested on data never seen during training, containing new users, locations, protocols, and devices.
        We select the datasets for the downstream tasks in order to have a diverse set of labels relevant for EDA data.
        We report in \cref{tab:simplified_datasets} the binary task from each dataset in the downstream evaluation part.

        \paragraph{Pre-processing}
        We pre-process the data from both \dataname-train and \dataname-test parts following related work on EDA data~\cite{Alchieri2024,DiLascio2019,gashiDetectionArtifactsAmbulatory2020,Hossain2022}. First, we apply a Butterworth low-pass filter (cutoff $0.4~\mathrm{Hz}$) to remove high-frequency noise. Then, we decompose the EDA signals into phasic and tonic components, using the cvxEDA method from~\citet{Greco2016}.
        This is a standard procedure when working with EDA data.
        The phasic component is associated with short-term changes, e.g., momentary stress, while the tonic component with longer-term variations~\cite{Boucsein2012}.
        To train the \modelname foundation model, we use both the phasic and tonic components, as well as the non-decomposed EDA signal.
        Using all three signals, i.e., phasic, tonic and original EDA signal, is a common procedure in classification-based tasks using EDA data, since it allows models to learn both short and long-term effects in the data~\cite{Alchieri2024,Alchieri2026}.

        \paragraph{Data segmentation}
        After pre-processing the data, we segment the EDA signals into fixed-length windows of 60 seconds. We use the same window length as \citet{Matton2023} and \citet{Schmidt2018}, who highlight how this length allows one to capture both short- and long-term changes in the EDA data.
        To train \modelname, on \dataname-train we select maximally overlapping windows, i.e., with an overlap step of $0.25~\mathrm{s}$ (the sampling period).
        We implement this approach as done by \citet{Matton2023}, on self-supervised learning for EDA data, and \citet{Ansari2024,Goswami2024,Feofanov2025}, on generalist time series foundation models for other physiological signals.
        With this approach, we obtain a training set consisting of approximately \emph{275 million} windows of EDA data.

        We also apply the same 60-second segmentation to the data used for evaluation, i.e., \dataname-test. For evaluation, we assign to each window a binary label, corresponding to the associated task: we refer to \cref{tab:simplified_datasets} for an overview of the binary downstream tasks used.
        Whenever a dataset contains EDA collected simultaneously from both sides of the body, for evaluation we use only the data from the body side most associated with the task, following guidelines by \citet{Alchieri2024}, e.g., right-side EDA signals for cognitive load classification.
        On \dataname-test, we use non-overlapping windows, since the literature shows that testing on either overlapping or non-overlapping windows leads to similar results~\cite{Dehghani2019,Tello2024}.

        \paragraph{Discussion on rescaling the data}
        We do not apply any normalization or standardization, e.g., min-max normalization, on the prepared windows of EDA data. Researchers working with EDA data collected in lab-controlled environments apply per-user min-max normalization to reduce inter-personal variability and improve classification performance significantly, e.g.,~\cite{Mishra2020,Schmidt2018,Almadhor2023}.
        However, EDA data is affected by both inter-personal variability, which min-max normalization addresses, and intra-personal variability, i.e., a user's data changes over time. A fixed calibration, like the one applied by min-max normalization, is rendered invalid over time if a user's signal morphology changes~\cite{Viana-Matesanz2024,Beten2025}.
        Moreover, ``cold-start'' approaches, i.e., where a machine learning model is applied directly to a new user, are preferred for wearable physiological data~\cite{Yi2023}, and in these scenarios normalization cannot be directly implemented.

        \begin{table}[t]
            \centering
            \caption{Summary of Datasets and Associated Binary Tasks.}
            \label{tab:simplified_datasets}
            \begin{tabular}{lll}
                \toprule
                \textbf{Dataset} & \textbf{Binary Tasks} & \textbf{Device} \\
                \midrule
                APSync & Low/High Engagement & Empatica E4 \\
                HeartS & Sleep/Wake & Empatica E4 \\
                USILaughs & Cog. Load/Relaxation & Empatica E4 \\
                WESAD & High/Low Arousal, High/Low Valence (self-report) & Empatica E4 \\
                DREAMT & Sleep/Wake, Deep Sleep/REM & Empatica E4 \\
                HHISS & Low/High Stress & Empatica E4 \\
                Amigos~\cite{miranda2018amigos} & Arousal High/Low, Liking High/Low & Shimmer 2R \\
                EDABE~\cite{llanes2023edabe} & Artifact & Shimmer3 GSR+ \\
                Multi-physio~\cite{lucas2026physiological} & Relax/Video & Shimmer3 GSR+ \\
                \bottomrule
            \end{tabular}
        \end{table}

    \subsection{Model Training Task \& Architecture}
    \label{sec:model_arch}

        \paragraph{Training objective}
        We adopt contrastive learning to train \modelname. Contrastive learning is used by multiple researchers to train foundation models for wearable physiological data, e.g., \cite{Abbaspourazad2024,Pillai2025}. This approach has also been used for generalist time series foundation models specialized in classification tasks, e.g., \cite{Feofanov2025}.
        Contrastive learning is a self-supervised learning method that consists in training an encoder-only model to learn a latent embedding space where representations of similar data pairs --- typically created via augmentations of the same signal --- are attracted to each other, while representations of dissimilar pairs are repelled.
        The latent embeddings often contain information which allows them to be used effectively in downstream tasks~\cite{Le-Khac2020}.
        We also experiment with an additional self-supervised learning method: masked reconstruction. This method consists in training an encoder-decoder to reconstruct the whole signal from a masked version of it, i.e., a signal which is missing some parts.
        We report in \cref{app:ablation} details about this additional experiment which, however, failed to learn useful representations of our EDA data.

        \paragraph{Model architecture}
        We choose the architecture for our foundation model from similar work in the literature~\cite{Abbaspourazad2024,Narayanswamy2024,Pillai2025,Feofanov2025}. Our choice reflects the decision to implement a generalist time series foundation model on EDA data. This choice is in line with research on the first PPG and ECG-specific foundation models~\cite{Abbaspourazad2024,Narayanswamy2024}: our objective is to showcase how features computed from EDA-specific foundation models perform, as well as to provide weights and code to the research community.
        Each physiological signal, from PPG to ECG and EDA, has morphological elements specific to it. It is possible to define architectures for foundation models that adapt to these morphological elements, as recent work from \citet{Saha2025} shows. However, this implementation goes beyond the scope of the current work.

        We chose an EfficientNet~\cite{Tan2019} architecture, as did \citet{Abbaspourazad2024}, since it is computationally less expensive than traditional convolutional neural network (CNN) backbones. We consider this choice also in light of the fact that foundation models for wearable data have the potential to be used on wearable devices themselves~\cite{Abbaspourazad2024}.
        We adapt the EfficientNet for our input data, i.e., 1-dimensional time series data of 240 values ($60\mathrm{s}$ at $4~\mathrm{Hz}$) with 3 channels (tonic, phasic and original EDA signal).
        We report in \cref{app:ablation} ablation studies on the model size and hyperparameters. We implement a version of EfficientNet with approximately 1M parameters and a latent representation of $d=64$.

    \subsection{Training Process}
    \label{sec:training_info}

        \paragraph{Train loss}
        Using the contrastive learning training objective and EfficientNet architecture defined in \cref{sec:model_arch}, we train \modelname using the InfoNCE loss~\cite{Oord2019}. Formally, for a given pair of embeddings $(z_i, z_j)$, the loss is defined as:
        \begin{equation}
        \label{eq:infonce_loss}
            \mathcal{L}_{i,j} = -\log \frac{\exp(\text{sim}(z_i, z_j) / \tau)}{\sum_{k=1}^{2N} \mathbb{1}_{[k \neq i]} \exp(\text{sim}(z_i, z_k) / \tau)}
        \end{equation}
        where $\text{sim}(\cdot, \cdot)$ denotes the cosine similarity, $\tau=0.1$ is a temperature hyperparameter, $N$ is the batch size, and $\mathbb{1}$ is the indicator function.
        This objective maximizes the similarity between representations of the same underlying signal while minimizing agreement with unrelated samples.

        \paragraph{Pairs of EDA data}
        We generate positive pairs of EDA signals by applying two distinct stochastic augmentations to the same EDA signal segment (an anchor segment). We employ the set of data augmentations optimized for EDA signals proposed by~\citet{Matton2023}. This set combines both standard data augmentation techniques (e.g., signal warping) and augmentations specific to EDA data (e.g., loose sensor artifact). Negative pairs consist of comparisons between the anchor segment and all other segments in the mini-batch.
        In \cref{app:user-wise pairs}, we report results of experiments conducted using user-assignment to generate positive pairs.
        We further report the full list of augmentations and additional details about the training in \cref{app:meta} (\cref{tab:eda_augmentations}).

    \section{Evaluation of \modelname}
    \label{sec:evaluation}
        In this section, we report the results of the evaluation procedure of \modelname on the selected downstream tasks from \dataname-test.
        We use \emph{linear probing} with frozen weights to evaluate the performance on the selected classification tasks, as frequently done to evaluate the performance of foundation models~\cite{Abbaspourazad2024,Saha2025,Pillai2025,Zhang2025}.
        We compare the \modelname feature set to various baseline features, including generic handcrafted features, a set of EDA-specific features, and features computed from generalist time series foundation models.
        Finally, we also evaluate the computational complexity of the \modelname foundation model in extracting features, compared with the other baseline methods.

        \subsection{Experimental Evaluation Setup}
            \paragraph{Evaluation through linear probing}
            We evaluate the features computed from the \modelname foundation model using \emph{linear probing} on the downstream tasks from \dataname-test.
            \label{c2.2 - 1}
            For each evaluation experiment, we use the foundation models, including \modelname, as feature extractors. To do this, we simply freeze, i.e., do not update during training, the foundation model weights when we pass the EDA data as input.
            We then train only a logistic regression classifier on the features computed using the foundation models.
            The usage of frozen weights and linear probing is commonly done in similar work on foundation models for physiological data~\cite{Abbaspourazad2024,Pillai2025,Saha2025}.
            We use the same evaluation procedure for \modelname and the other baseline feature sets.
            We use a linear model for all approaches since we are gauging the representativeness of the feature sets on classification tasks, and not that of the downstream model itself.
            Linear probing also allows us to estimate the linear separability of the different feature sets.

            \paragraph{Cross-validation protocols for the downstream tasks}
            We evaluate the linear probing using two distinct cross-validation methods. The first method is Leave-One-Participant-Out (LOPO) cross-validation. LOPO cross-validation is used by researchers to evaluate how machine learning models generalize to new users~\cite{Rehman2024}. With this method, we test robustness to inter-personal variability.
            The second validation method is Time-Aware (TA) cross-validation~\cite{Alchieri2025}, which we use to evaluate the model's ability to generalize to data from users already seen in the training set, testing robustness to intra-personal variability. We partition users into $N=5$ folds: the model trains on all external groups plus the first chronological 2/3 of the target fold, reserving the final 1/3 of data strictly for testing. This approach simulates a realistic scenario where a model uses a specific participant's historical data to predict their future states.

            In both validation methods, we perform hyperparameter tuning for the logistic regression at each cross-validation iteration. We select the hyperparameters using a 3-fold inner cross-validation with grid search.
            We report in \cref{app:meta} information about the hyperparameter grid used.

            \paragraph{Baseline feature sets}
            \label{par:baseline_features}
            We compare the performance using linear probing on the features computed from \modelname and using other baseline methods.
            Specifically, following the same experimental setup as similar work on foundation models for physiological data~\cite{Abbaspourazad2024,Pillai2025,Narayanswamy2024}, we define a set of baseline, \emph{generic}, handcrafted features. These features are: the mean, the standard deviation, the minimum and the maximum of a $60\mathrm{s}$ EDA signal. We compute these features for all three EDA components, i.e., phasic, tonic and original signal.
            In total, the dimensionality of this feature set is $d=12$.

            However, the aforementioned generic handcrafted feature set does not represent the state-of-the-art approach for EDA-based classification tasks~\cite{Alchieri2024,gashiDetectionArtifactsAmbulatory2020,Lutin2021}. To this end, we also implement a second handcrafted feature set, which we call \emph{EDA-specific} handcrafted features. This feature set includes both statistics, e.g., average of the first derivative, and features specific to EDA data, e.g., number of EDA peaks and their average amplitude.
            As with the generic handcrafted feature set, we compute these features on $60\mathrm{s}$ windows and for all three EDA components.
            In total, the dimensionality of the EDA-specific handcrafted feature set is $d=45$.

            We also compare the performance using linear probing of the feature set from \modelname with other generalist time series foundation models. We compare with the following: Chronos~\cite{Ansari2024}, MOMENT~\cite{Goswami2024} and Mantis~\cite{Feofanov2025}. We select these three foundation models since \citet{Alchieri2026} show that they achieve performance similar to the EDA-specific handcrafted features when using EDA data.
            \citet{Saha2025} also show that Chronos~\cite{Ansari2024} and MOMENT~\cite{Goswami2024} achieve, on average, performance similar to their PPG-specific foundation model on the downstream tasks selected by the authors.
            We also select the recent foundation model Mantis~\cite{Feofanov2025} since it is trained for classification tasks on time series data.

            Mantis takes time series of length 512 as input: we oversample our time series, which have a length of 240, to match this desired length.
            From a single time series, Mantis computes a set of embeddings, similarly to our foundation model.
            Mantis has a feature set size of $d=768$ (embedding size of $256$ across three channels). The embedding size of MOMENT is $d=1024$.
            However, Chronos computes embeddings ($d=512$) for each timepoint in the series. This leads, with our input, to high dimensionality of the data. To address the high dimensionality problem, we average across the time axis the embeddings extracted from each timepoint. Overall, the feature set from Chronos is of size $d=1536$ (embedding size of $512$ across three channels).

            \paragraph{Fine-tuning \modelname and Mantis}
            \label{par:fine-tune}
            In addition to the linear probing validation methodology, we implement a fine-tuning-based validation. In this scenario, instead of freezing the foundation model weights, and consequently using the models as feature extractors, we fine-tune the foundation models on the downstream data. We implement this validation method for \modelname and Mantis.
            Our fine-tuning protocol follows that used by \citet{Feofanov2025} and \citet{Saha2025}. We attach a linear head to the foundation models and train the model together with the new head.
            We fine-tune on each downstream task for up to 50 epochs, with early stopping on the validation macro F1-score, at each cross-validation iteration.
            We provide additional details about the fine-tuning protocol and the chosen hyperparameters in \cref{app:finetuning}.

            \paragraph{Reporting and task selection information}
            We report all results in terms of \emph{balanced accuracy}. We use this metric since the downstream evaluation datasets we use contain binary labels which, in a subset of cases, are imbalanced. Balanced accuracy accounts for class imbalance, reporting results which are more representative of the real performance on a specific task~\cite{Brodersen2010,Owusu-Adjei2023}. At the same time, compared to other metrics for imbalanced data, e.g., Matthews correlation coefficient (MCC), balanced accuracy is more interpretable~\cite{Grandini2020}.

            In addition to the results from the identified feature sets, we also report results from a dummy classifier. The dummy classifier reported is the one achieving the highest balanced accuracy in each given task, among the following: \emph{most frequent}, which always predicts the most frequent class; \emph{uniform}, which predicts the binary labels by drawing randomly from a uniform distribution; and \emph{prior}, which predicts the binary labels by drawing randomly from the distribution of labels in the training set.

            We report in this section results from tasks that are \emph{solvable}, i.e., at least one feature set achieves balanced accuracy higher than the dummy classifier. If no model achieves balanced accuracy higher than, for example, random chance, we conclude that it is not due to issues with the feature sets, but with the task itself, e.g., the task is not solvable with the given constraints.
            In \cref{app:additional_results}, we also report MCC and F1-score results, including for tasks not considered in \cref{tab:binary_results_TA_lopo}, for which the linear classifier did not significantly outperform the dummy classifier for any feature set.

        \subsection{Comparison with Generic Handcrafted Feature Set}

\begin{figure}[t]
              \centering
              \includegraphics[width=0.7\linewidth]{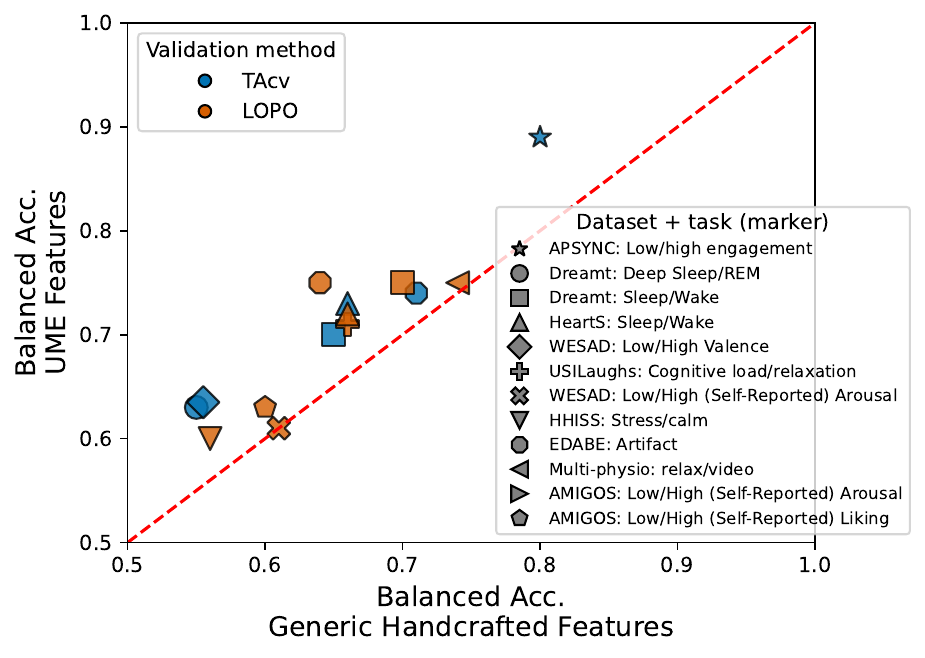}
              \caption{Pairplot with comparison between \modelname and the baseline features for the selected downstream tasks. Colors represent the two validation methods, TA and LOPO.}
              \Description{Pairplot with comparison between \modelname and the baseline features for the selected downstream tasks.}
              \label{fig:pairplot_main}
            \end{figure}

            We present in \cref{fig:pairplot_main} the comparison when performing linear probing on the selected downstream tasks, between the feature set from our \modelname foundation model and the generic handcrafted features.
            We show results for both validation methods, i.e., LOPO and TA.
            The results show that the features from our foundation model outperform the generic handcrafted features in 14 out of the 15 experiments, including datasets containing non-Empatica E4 EDA data.
            The improvement holds true regardless of the scenario and the task selected.
            We conclude from these findings that the \modelname foundation model learns features useful for EDA data in performing downstream predictions.
            Moreover, the learned representations are not specific to the EDA from Empatica E4 devices, but translate to different devices (Shimmer 2R and 3 GSR+) as well.
            Our findings are in line with work on foundation models for PPG, when comparing to generic handcrafted feature sets~\cite{Abbaspourazad2024,Pillai2025,Narayanswamy2024}.

        \subsection{Comparison with Other Baseline Feature Sets}
        \label{sec:comparison_complete}

            We report in \cref{tab:binary_results_TA_lopo} the results for the evaluation of our \modelname foundation model, in terms of balanced accuracy and for the two validation methods introduced above, i.e., TA and LOPO cross-validation.
            We compare the feature set computed using \modelname to the following feature sets: a set of generic handcrafted features; a set of EDA-specific handcrafted features; and features computed using generalist time series foundation models, specifically Mantis~\cite{Feofanov2025}, Chronos~\cite{Ansari2024} and MOMENT~\cite{Goswami2024}.

            \paragraph{Results using TA} The TA cross-validation method tests the intra-personal generalizability of models trained using the different feature sets, i.e., the ability to generalize to new data from users already seen in the training data.
            Our first finding is that, using linear probing, models trained using features from \modelname outperform or match generic and EDA-specific handcrafted features. We conclude that the embeddings from \modelname capture user-specific dynamics, which allow us to obtain a balanced accuracy on par with, or higher than, handcrafted-based methods.
            We also find that models trained using the feature set from Mantis~\cite{Feofanov2025}, which is trained to solve generic classification tasks, perform similarly to those trained using the feature set from \modelname. We notice that the other two foundation models, MOMENT~\cite{Goswami2024} and Chronos~\cite{Ansari2024}, have lower performance, on a majority of tasks, than both the two handcrafted feature sets and the foundation models (\modelname and Mantis~\cite{Feofanov2025}).

            \paragraph{Results using LOPO} The LOPO cross-validation method tests the inter-personal generalizability of models trained using the different feature sets, i.e., the ability to generalize to data from new users not seen in the training set.
            We find that models trained from the \modelname features have similar balanced accuracy to models trained from either the EDA-specific handcrafted features or the embeddings computed using Mantis~\cite{Feofanov2025}. Features from the other foundation models have similar or lower performance as well.
            Generalization to new users is a known issue when working with EDA data~\cite{gashiDetectionArtifactsAmbulatory2020}. While features from foundation models, both our model and others, achieve performance similar to that of EDA-specific handcrafted features, they do not solve the ``cold-start'' problem.

            \paragraph{Performance on unseen devices}

            We also test downstream classification performance on unseen devices. \modelname outperforms both handcrafted feature sets in four of these five experiments, with similar overall performance compared to when testing on data from E4 devices.
            Our results suggest that \modelname generalizes to EDA data from two unseen wearable devices, with no substantial degradation with respect to the device it was pretrained on.
            However, device, dataset, task, participant population, and collection protocol vary together across these datasets. Further analysis should evaluate on unseen devices while controlling for these other factors.

            \paragraph{Results using fine-tuning}
            In \cref{tab:binary_results_TA_lopo} we also report the results obtained by fine-tuning \modelname and Mantis~\cite{Feofanov2025} on each downstream task, instead of using them as frozen feature extractors.
            Fine-tuning converged in 14 of the 15 experiments: on APSync neither model converged, which we report as FTC in \cref{tab:binary_results_TA_lopo}, and we exclude that experiment from the counts below and from the statistical analysis.
            When comparing the two fine-tuned models, \modelname achieves a higher balanced accuracy than Mantis in 11 of these 14 experiments, with one tie and two experiments in which Mantis is higher.
            Taking the frozen and fine-tuned variants together, \modelname obtains the best or tied-best result in 10 of the 15 experiments.
            Importantly, fine-tuning never degrades \modelname's performance beyond the standard error of the corresponding linear probing result, and improves it in 10 of the 14 experiments, leaving it unchanged in two, with the largest gains on the tasks where the frozen features perform worst, e.g., low/high arousal in WESAD ($.61 \rightarrow .70$). Fine-tuning thus allows \modelname to better adapt to the downstream tasks.

            \paragraph{Result analysis using Friedman-Nemenyi test}
            We perform the Friedman test, followed by the post-hoc Nemenyi test, to compare model performance across all experiments~\cite{Demsar2006}. We consider each combination of dataset-validation method as a single sample for the statistical analysis.
            Overall, the Friedman test ranks fine-tuned \modelname as the model with the highest performance, with the non-fine-tuned Mantis and \modelname following behind with a rank score similar to each other. We refer to \cref{tab:friedman-ranks} in \cref{app:additional_results} for the full ranking list.
            When testing for statistical significance using the Nemenyi test, we find that \modelname, without fine-tuning, outperforms the generic feature set ($p=0.0495 < \alpha = 0.05$). However, when fine-tuned, \modelname significantly outperforms both the generic feature set ($p = 5.67 \times 10^{-5} < \alpha = 0.05$) and the EDA-specific features ($p=0.0401 < \alpha =0.05$).

            From these findings we conclude that \modelname and Mantis~\cite{Feofanov2025}, without fine-tuning, capture time-series dynamics that are relevant for EDA classification. When specializing the models on the downstream tasks, through fine-tuning, we find that \modelname achieves the overall highest rank, while Mantis actually underperforms compared to its non-fine-tuned version. We think this is the case due to the small size of the downstream datasets, compared to the larger parameter count of Mantis.

            \begin{table}[t]
                \centering
                \footnotesize
                \caption{Results of binary classification experiments across datasets and tasks, in terms of \emph{balanced accuracy}$_{standard~error}$. Reported are results for both TA and LOPO cross-validation methods. Acronyms: \emph{Gen. HC} stands for \emph{generic handcrafted features}; \emph{EDA HC} stands for \emph{EDA-specific handcrafted features}; FTC stands for \emph{failed to converge}; in the \emph{Device} column, \emph{E4} stands for \emph{Empatica E4}, \emph{S3} for \emph{Shimmer3 GSR+}, and \emph{S2} for \emph{Shimmer 2R}. Our \modelname is pretrained using Empatica E4 data only.}
                \label{tab:binary_results_TA_lopo}
                \renewcommand{\arraystretch}{1.3}
                \setlength{\tabcolsep}{3pt}
                \begin{tabularx}{\textwidth}{l c X | c | c c | c c c | c | c c}
                    \toprule
                    & & & \multirow{2}{*}{\textbf{Dummy}} & \multicolumn{2}{c|}{\textbf{HC}} & \multicolumn{3}{c|}{\textbf{Generalist Foundation}} & \textbf{Ours}  & \multicolumn{2}{c}{\textbf{Fine tune}}\\
                    \textbf{Dataset} & \textbf{Device} & \textbf{Binary Task}
                    &
                    & {Gen.}
                    & {EDA-spec.}
                    & {Mantis}
                    & {MOMENT}
                    & {Chronos}
                    & {\modelname}
                    & {\modelname}
                    & Mantis
                    \\[0.5ex]
                    \cline{4-12}
                    & & & \multicolumn{9}{c}{\emph{Balanced accuracy}$_{standard~error}$} \\[0.5ex]

                    \rowcolor{gray!12}
                    \multicolumn{12}{l}{\textbf{\scshape Time-Aware cross-validation (TA)}} \\
                    \addlinespace[3pt]

                    APSync & E4 & Low/High engagement
                    & $.45_{.20}$ & $.80_{.10}$ & $.80_{.10}$ & $.86_{.07}$ & $.47_{.16}$ & $.61_{.20}$ & $.89_{.11}$ & FTC & FTC \\

                    \addlinespace
                    DREAMT & E4 & Deep Sleep/REM
                    & $.48_{.02}$ & $.55_{.05}$ & $.59_{.04}$ & $.64_{.03}$ & $.65_{.01}$ & $.63_{.02}$ & $.63_{.05}$ & $.67_{.03}$ & $.65_{.04}$ \\

                    DREAMT & E4 & Sleep/Wake
                    & $.50_{.01}$ & $.65_{.02}$ & $.69_{.01}$ & $.73_{.01}$ & $.69_{.01}$ & $.73_{.01}$ & $.70_{.01}$ & $.74_{.01}$ & $.65_{.01}$ \\

                    \addlinespace
                    HeartS & E4 & Sleep/Wake
                    & $.49_{.00}$ & $.66_{.04}$ & $.72_{.07}$ & $.75_{.09}$ & $.69_{.06}$ & $.74_{.06}$ & $.73_{.09}$ & $.71_{.06}$ & $.72_{.05}$ \\

                    \addlinespace
                    WESAD & E4 & Low/High Valence
                    & $.51_{.03}$ & $.55_{.10}$ & $.54_{.10}$ & $.61_{.04}$ & $.64_{.05}$ & $.45_{.06}$ & $.63_{.06}$ & $.69_{.04}$ & $.61_{.03}$\\

                    EDABE & \shimmer{S3} & Artifact
                    & $.49_{.01}$ & $.71_{.03}$ &  $.72_{.02}$ & $.77_{.05}$ & $.73_{.02}$ & $.77_{.05}$ & $.74_{.03}$ & $.81_{.01}$ & $.76_{.02}$ \\

                    \addlinespace[6pt]
                    \rowcolor{gray!12}
                    \multicolumn{12}{l}{\textbf{\scshape Leave-one-participant-out (LOPO) cross-validation}} \\
                    \addlinespace[3pt]

                    DREAMT & E4 & Sleep/Wake
                    & $.48_{.00}$ & $.70_{.01}$ & $.74_{.01}$ & $.76_{.01}$ & $.73_{.01}$ & $.78_{.01}$ & $.75_{.01}$ & $.76_{.01}$ & $.75_{.01}$ \\

                    \addlinespace
                    USILaughs & E4 & Cog. load/relax
                    & $.50_{.00}$ & $.66_{.03}$ & $.70_{.04}$ & $.72_{.05}$ & $.60_{.05}$ & $.67_{.03}$ & $.71_{.05}$ & $.74_{.04}$ & $.72_{.04}$ \\

                    \addlinespace
                    HeartS & E4 & Sleep/Wake
                    & $.50_{.00}$ & $.66_{.04}$ & $.70_{.03}$ & $.74_{.03}$ & $.70_{.02}$ & $.73_{.03}$ & $.72_{.03}$ & $.71_{.04}$ & $.71_{.05}$ \\

                    \addlinespace
                    WESAD & E4 & Low/High Arousal
                    & $.58_{.05}$ & $.61_{.04}$ & $.63_{.04}$ & $.56_{.04}$ & $.66_{.04}$ & $.66_{.05}$ & $.61_{.05}$ &  $.70_{.04}$  &  $.61_{.05}$  \\

                    \addlinespace
                    HHISS & E4 & Stress/calm
                    & $.50_{.00}$ & $.56_{.02}$ & $.64_{.01}$ & $.63_{.02}$ & $.55_{.01}$ & $.59_{.01}$ & $.60_{.02}$ &  $.60_{.02}$ & $.56_{.01}$ \\

                    \addlinespace

                    EDABE & \shimmer{S3} & Artifact
                    & $.50_{.00}$ & $.64_{.01}$ & $.68_{.01}$ & $.72_{.01}$ & $.70_{.01}$ & $.75_{.02}$ & $.69_{.01}$ & $.75_{.01}$ & $.72_{.01}$ \\
                    Multi-physio & \shimmer{S3} & relax/video
                    & $.49_{.02}$ & $.74_{.07}$ & $.79_{.07}$ & $.74_{.06}$ & $.69_{.06}$ & $.56_{.06}$ & $.75_{.06}$ & $.75_{.06}$ & $.81_{.06}$ \\

                    Amigos & \shimmer{S2} & arousal high/low
                    & $.50_{.00}$ & $.46_{.03}$ & $.48_{.03}$ & $.55_{.04}$ & $.49_{.01}$ & $.50_{.03}$ & $.60_{.04}$ & $.61_{.06}$ & $.49_{.02}$ \\
                    Amigos & \shimmer{S2} & liking high/low
                    & $.54_{.03}$ & $.60_{.05}$ & $.58_{.06}$ & $.61_{.04}$ & $.57_{.03}$ & $.53_{.04}$ & $.63_{.05}$ & $.66_{.12}$ & $.58_{.04}$ \\
                    \bottomrule
                \end{tabularx}
            \end{table}

    \subsection{Computational Complexity Analysis}
        We compare the computational complexity of the different feature extraction methods, i.e., the handcrafted-based approaches, our \modelname foundation model, and the other baseline foundation models. We compare the methods using both \emph{CPU execution time} and the number of floating-point operations (FLOPs)~\cite{Chen2023}. We choose this dual approach since handcrafted features cannot be compared using FLOPs alone.
        We perform all calculations using an Apple M1 Max CPU.

        \begin{figure}[t]
            \centering
            \begin{subfigure}[b]{0.48\linewidth}
                \centering
                \includegraphics[width=\linewidth]{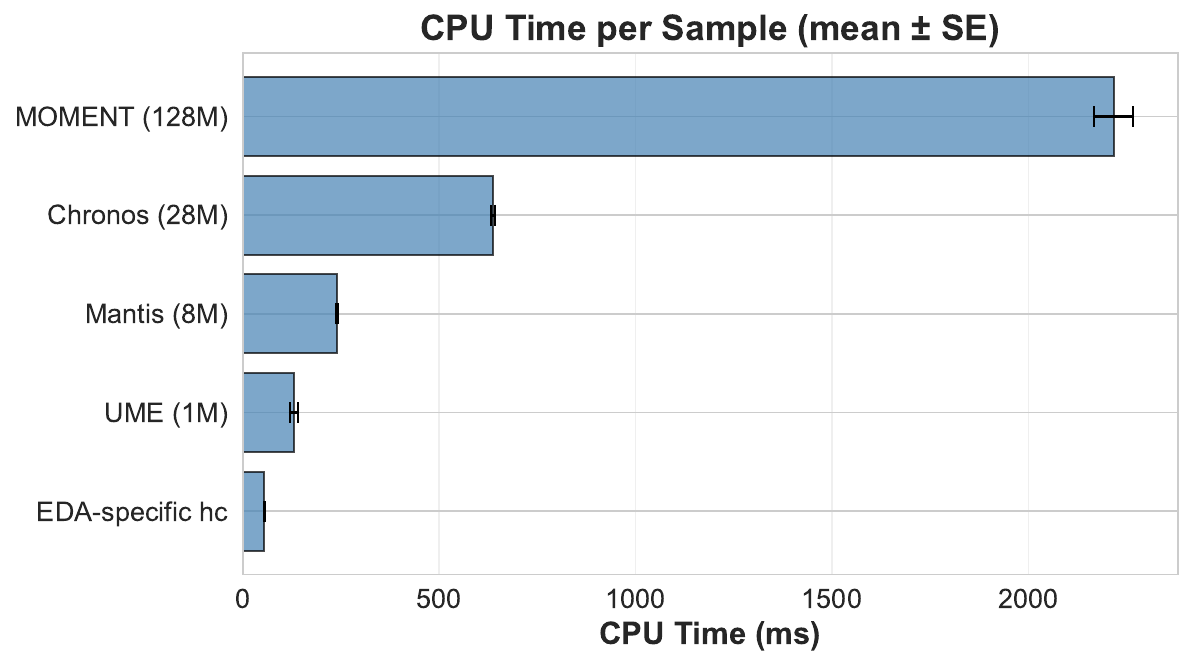}
                \caption{CPU Time}
                \label{fig:computational_complexity_cpu}
            \end{subfigure}
            \hfill
            \begin{subfigure}[b]{0.48\linewidth}
                \centering
                \includegraphics[width=\linewidth]{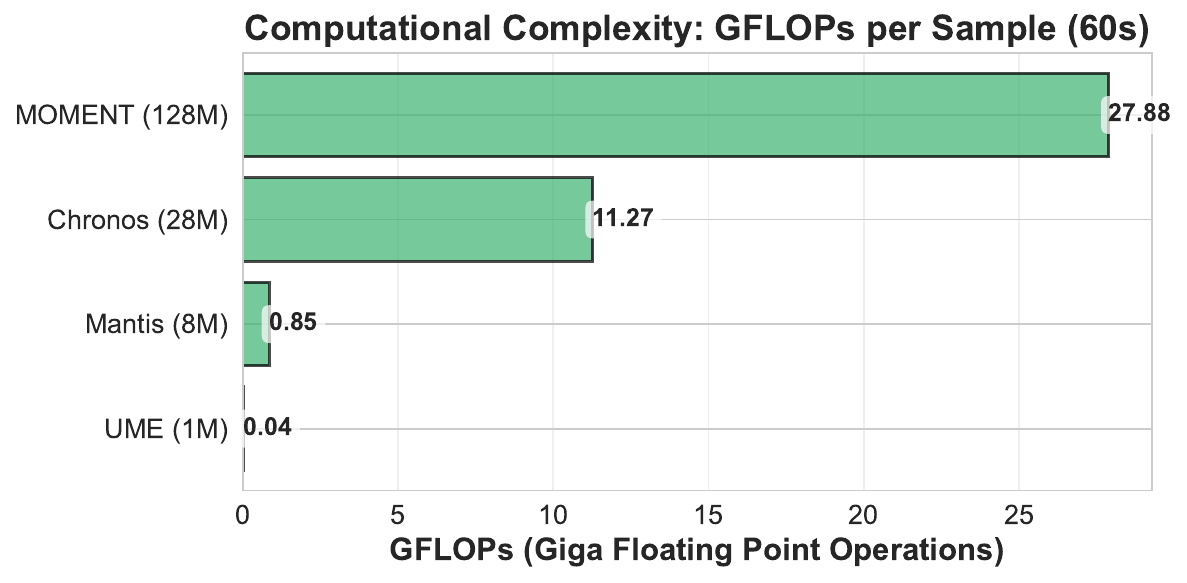}
                \caption{Model GFLOPs}
                \Description{Model GFLOPs}
                \label{fig:computational_complexity_placeholder}
            \end{subfigure}
            \caption{Computational complexity analysis. \textbf{(a)} Average (across 20 samples) per-sample computation time for a 60-second window of EDA data for all three components (phasic, tonic, and original signal). Note: \emph{hc} stands for \emph{handcrafted}; generic handcrafted features are not shown as they require negligible computation time. \textbf{(b)} GFLOPs for the different foundation models.}
            \label{fig:computational_complexity}
        \end{figure}

        To get \emph{CPU execution time}, we use a random subset of twenty $60\mathrm{s}$ windows from the downstream part of the \dataname collection. We additionally use three samples per experiment as warm-up runs.
        We perform FLOPs computation using the \texttt{fvcore} Python library.
        We report in \cref{fig:computational_complexity} the results. We find that the handcrafted feature extraction methods have the lowest CPU execution time, as expected.
        We also find that our \modelname foundation model is significantly faster than all other foundation models, both with respect to \emph{CPU execution time} and FLOPs.
        We conclude that our model, which is trained only on EDA data, achieves performance on par with larger, general-purpose foundation models on a set of downstream tasks using EDA data, as detailed in \cref{sec:comparison_complete}. However, it does so with a fraction of the computational resources needed, while also eliminating the need for expert-designed or handcrafted features, unlike traditional feature-based pipelines.

\section{Discussion, Limitations and Future Work}
\label{sec:discussion}

\paragraph{Summary of results}
Through this paper, we provide the research community with the \dataname collection of datasets, which enabled the development of \modelname, the first open-source foundation model for wearable EDA. We find that \modelname, trained on a subset of the \dataname collection using contrastive learning, surpasses the performance obtainable using generic handcrafted features and is comparable to, or better than, EDA-specific handcrafted features.
Further, \modelname matches the performance of larger and much less computationally efficient generalist foundation models for time series data, when using either a ``frozen backbone'' or fine-tuning approach.
\modelname additionally generalizes to EDA data from two devices unseen during pretraining.
By relying on publicly available data only and by releasing all artifacts produced as part of this work, we aim to foster further research on foundation models for wearable EDA as well as other research related to unimodal and multimodal EDA sensing and modeling.

\paragraph{Discussion}
Overall, we observe that features computed using \modelname, generalist time-series foundation models, and EDA-specific handcrafted features lead to comparable downstream performance, even in cases where we fine-tune the foundation models.
It can be argued that other methodologies, in particular the use of domain-specific architectural elements, may yield superior performance.
However, we believe that our results are not bound to \modelname's specific characteristics, because \modelname's performance not only matches that of generalist foundation models, but also often surpasses results obtained relying on EDA-specific handcrafted features, with fine-tuning further improving performance. This suggests that other methodological approaches alone may yield incremental rather than substantial gains.

Instead, we believe that the reported results, obtained by \modelname and other methods, are close to the intrinsic upper limit reachable for the considered downstream tasks, due to the arguably limited information in the input signal that is predictive of the label.
Potential additional gains can however be obtained by coping with the significant label noise and signal noise of EDA data. Wearable EDA is noisy, strongly affected by motion/contact artifacts and low signal-to-noise ratio, in particular when collected in the wild~\cite{gashiDetectionArtifactsAmbulatory2020}. Additionally, high-level constructs (e.g., stress, engagement, valence) might only be weakly identifiable from EDA alone without additional context.
We believe that this represents an important direction for future work, e.g., by quantifying the impact of signal quality filtering, label reliability, and more.

A further element to highlight is that \modelname, which is specialized for EDA data, requires far fewer parameters to match the performance on downstream tasks of larger generalist models.
As highlighted by \citet{Abbaspourazad2024}, developing foundation models for wearable data includes considering the usage of these models on actual wearable devices, which are resource constrained. We speculate that our results showcase how specializing foundation models for EDA data limits their size and, consequently, opens up prospects for their usage on wearable devices.

\paragraph{Limitations \& future work}
\dataname contains data from three different EDA devices. However, we only use Empatica E4 data during the pretraining of our foundation model, due to the limited availability of non-Empatica datasets in the literature.
We show that the model transfers to three datasets recorded with unseen devices; however, device and dataset characteristics are confounded in this evaluation. We encourage the research community to gather EDA data from new sources and devices and to potentially pretrain EDA foundation models using multiple device sources as well as expand on our analysis.

\dataname-test contains nine datasets with binary classification downstream tasks. While they cover most tasks relevant for EDA, our evaluation setup does not include tasks other than binary classification. Future work should explore additional classification or regression tasks, e.g., multi-class emotion classification.

We also fine-tune on datasets ranging from about 1 to about 900 hours, but without systematically varying the amount of labeled data: a dedicated few-shot adaptation analysis remains future work.

In our pre-processing approach, we apply standard low-pass filtering and decomposition of the EDA signals using the cvxEDA method~\cite{Greco2016}. Alternative filtering or decomposition techniques~\cite{Veeranki2024} can be explored in future research.

Lastly, we acknowledge that the \dataname collection of datasets is affected by the same user and domain biases associated with the individual datasets.

\section{Conclusions}
\label{sec:conclusions}

In this work, we present \dataname, a large-scale collection of wearable EDA data created from \numdatasets datasets and comprising approximately 23{,}000 hours of data from 731 users. Leveraging the diversity and variability of \dataname, we train \modelname, the first foundation model for wearable EDA. Additional analyses of foundation model scaling laws confirm that the size of \dataname is well suited for training a specialized model.
We release \dataname, under a tiered structure, to the research community to support the development of future EDA foundation models.
Our experiments show that \modelname's latent representations outperform or match both handcrafted-based methods and generalist foundation models. Additionally, \modelname generalizes to EDA data collected from two devices different from the Empatica E4 used to train the model. When fine-tuned, \modelname achieves the overall highest model rank, indicating that it can further specialize to the downstream applications successfully. Our findings on EDA are consistent with prior work on other physiological signals~\cite{Abbaspourazad2024,Pillai2025}. At the same time, compared to generalist models such as Chronos~\cite{Ansari2024} and Mantis~\cite{Feofanov2025}, \modelname is at least $20\times$ more computationally efficient (in terms of GFLOPs), supporting deployment on real-world wearable devices, in line with recent work on foundation models for wearable devices~\cite{leeHiMAEHierarchicalMasked2025,zhouPhysiologyAwareMaskedCrossModal2026}.
We hope that this work supports further advances in EDA research by providing both the \dataname dataset collection and the open‑source \modelname foundation model with publicly available weights.

\section{Dataset and Code Availability}
\label{sec:dataset_code_avail}

We make the \dataname dataset collection available to the research community, in a tiered system, which we summarize in \cref{tab:access_tiers}. Of the \numdatasets datasets included in \dataname, 16 are re-shared under their original open-source licenses after being prepared in a unified format. The other 11 datasets, which we accessed under individual data-sharing agreements, are re-shared with permission from their respective authors under a single data-sharing agreement. For the remaining two datasets, DREAMT and HHISS, which we cannot re-share due to restrictions in their original licenses, but which are freely available from their original providers (\cref{app:datasets_info}), we provide the code needed to process the raw data and convert it into the same format used for the other datasets in \dataname. \cref{tab:appendix_datasets} in \cref{app:meta} provides an overview of the licenses associated with all datasets included in \dataname.

We share the following artifacts:
\begin{itemize}
    \item A website containing information about how to access \dataname, with a link to the open-source version and the data sharing agreement for the restricted version: \\ \url{https://pc.inf.usi.ch/datasets/edamame/};
    \item The Python code we used to report all data traces in a unified format and to pre-process the data for our analysis: \\ \url{https://github.com/LeonardoAlchieri/EDAMAME};
    \item The weights and the Python code needed to reproduce the pretraining of \modelname: \\ \url{https://github.com/LeonardoAlchieri/eda-foundation-models};
    \item The Python code we developed to perform downstream evaluations: \\ \url{https://github.com/USI-PCC/EDAMAME-downstream-evaluation}.
\end{itemize}

\paragraph{Ethics and data governance}
\label{par:ethics}
    \dataname aggregates and, in part, redistributes anonymized EDA data originating from human participants. Each constituent dataset was collected by its original authors under their own institutional ethics approvals; \dataname constitutes a secondary use of data that is either publicly released or shared under agreement. Our institution's delegate for ethics determined that secondary analysis of these already-collected, de-identified datasets did not require additional review. We re-share data only under its original open license or, where obtained under agreement, with the explicit permission of the original authors and under the same conditions. \dataname carries only pseudonymous participant identifiers, which we newly remap within \dataname, and limited demographic and contextual data. We retain timestamps and contextual metadata as released in the source datasets; these could permit linkage to the source datasets, but add no information beyond what their providers already release; we consider the re-identification risk low, also because EDA does not support reliable person identification~\cite{piciucco2021biometric}. Downstream use is restricted to non-commercial research and excludes clinical use, except for the individual datasets whose original licenses allow it. We also restrict individual-level decisions about people, and any attempt to re-identify participants.

\begin{table}[t]
    \centering
    \small
    \caption{Access tiers for the EDAMAME collection. Datasets are grouped by the data-sharing conditions under which they can be obtained, with the total amount of EDA data and number of users in each tier. Per-dataset licenses are reported in \cref{tab:appendix_datasets}.}
    \label{tab:access_tiers}

    \begin{tabularx}{\textwidth}{l r r r X}
        \toprule
        \textbf{Access Tier} & \textbf{\# Datasets} & \textbf{Duration (h)} & \textbf{\# Users} & \textbf{Access condition} \\
        \midrule
        Open license (redistributed)
            & 16 & 4,911 & 328
            & None --- original open licenses \\
        \addlinespace
        Data-sharing agreement (re-shared)
            & 11 & 16,975 & 257
            & Single agreement covering all 11 \\
        \addlinespace
        Restricted (reconstructible)
            & 2 & 906 & 146
            & Download from original providers (\cref{app:datasets_info}) \\
        \bottomrule
    \end{tabularx}
\end{table}

\section*{Acknowledgments}
This work is partially supported by the Swiss National Science Foundation (SNSF) through the grant 205121\_197242 for the project ``PROSELF: Semi-automated Self-Tracking Systems to Improve Personal Productivity''.

During the preparation of this work, the authors used GenAI tools (Copilot, ChatGPT, Gemini, and Claude) to check the grammar of selected paragraphs, to re-format tables already created by the authors and to help with the template formatting. After using these tools, the authors reviewed and edited the content as needed and take full responsibility for the content of the manuscript.
\bibliographystyle{plainnat}
\bibliography{main}

\clearpage

\appendix
\crefalias{section}{appendix}
\crefalias{subsection}{appendix}
\crefalias{subsubsection}{appendix}
\counterwithin{table}{section}
\counterwithin{figure}{section}

\section{Methodology, Evaluation, Transparency, and Availability (META) Appendix}
\label{app:meta}

    \subsection{Additional Dataset Information}
    \label{app:datasets_info}
    We report in \cref{tab:appendix_datasets} information about the original license associated with the datasets making up the \dataname collection. The datasets that we re-share are all under their original license, if so required. For the datasets that require a data sharing agreement to be signed, we contacted the original authors, who all agreed to the re-sharing under the conditions of the original data sharing agreement.

    The restrictions in the original licenses of DREAMT~\cite{Wang} and HHISS~\cite{xiao2025hhiss} do not allow us to re-share them, but both are freely available from their original providers: DREAMT from PhysioNet, at \url{https://physionet.org/content/dreamt/2.2.0/}, and HHISS from GitHub, at \url{https://github.com/asalekin-ubiquitouslab/HHISS_IMWUT/tree/main/data}.

    \subsection{Baseline Feature Sets}
    \label{app:handcrafted_baselines}

    In \cref{par:baseline_features} we explain how we use two handcrafted feature sets as baselines to evaluate the embeddings from our \modelname foundation model. The first feature set, which we call \emph{generic handcrafted feature set}, consists of four statistics, i.e., mean, minimum, maximum and standard deviation, computed for both the phasic and tonic EDA components, as well as the original non-decomposed EDA signal. This feature set emulates the baseline used by related work to evaluate foundation models for physiological data~\cite{Abbaspourazad2024,Pillai2025}.
    The second set of features, which we call \emph{EDA-specific handcrafted feature set}, is a broader set of features commonly used in EDA classification pipelines~\cite{Alchieri2024,gashiDetectionArtifactsAmbulatory2020}. In addition to the generic statistics, this set incorporates signal dynamics (e.g., slopes and derivatives), morphological peak characteristics, and frequency-domain components. This set consists of 15 base features, resulting in a total of 45 features per window (15 features $\times$ 3 EDA components).

    \cref{tab:features_appendix} details the exact mathematical formulations for all 15 base features and specifies their inclusion in the respective baseline sets.

    We report in \cref{tab:model_sizes_dimensions} the feature size of the feature extraction methods used in this work.

    \cref{tab:model_sizes_dimensions} also reports the size, in number of parameters, of the generalist time series foundation models we use as baselines. Both Chronos~\cite{Ansari2024} and MOMENT~\cite{Goswami2024} are released in several sizes, and the number of parameters we report refers to the specific pretrained variant we use, i.e., the \emph{Small} variant of Chronos ($\sim$ 28 M parameters) and the \emph{Large} variant of MOMENT ($\sim$ 128 M parameters). For Mantis~\cite{Feofanov2025}, we use the publicly released model ($\sim$ 8 M parameters).
    We note that these counts refer to the \emph{encoder} only and are therefore smaller than the total parameter counts reported in the original papers.
    In initial exploratory studies, we also experimented with other released variants of both Chronos and MOMENT. We report results only for the variant of each model that worked best in our setting, and we use the same variant consistently across all downstream tasks, both validation methods, and the computational complexity analysis (\cref{tab:extractor-performance}).

    \subsection{Hyperparameter Grid for the Logistic Regression}

    In \cref{tab:logistic_regression_hyperparams} we report the hyperparameter grid we used with the logistic regression. We used a 3-fold inner cross-validation to search for the best configuration of hyperparameters during our feature evaluation through linear probing.

    \subsection{Training Details}
    We train the \modelname foundation model using the Adam optimizer~\cite{kingma2014adam} (learning rate 0.001 and weight decay 0.01). We also use a learning rate scheduler, Reduce On Plateau (factor 0.5), to decrease the learning rate during training. We train the model for a maximum of 400 epochs, with early stopping and a batch size of 512.
    In total, we train our model for approximately 5 days, using an Nvidia A6000 GPU\footnote{\url{https://www.nvidia.com/en-us/products/workstations/rtx-a6000/}}.
    We implement the foundation model in Python, using the PyTorch~\cite{Paszke2019} and PyTorch Lightning libraries.

    \subsection{Fine-Tuning Details}
    \label{app:finetuning}

    When fine-tuning, \modelname is used as an end-to-end classifier rather than as a frozen feature extractor: the pretrained weights serve as initialization, a linear head is attached on top of the $d=64$ embedding, and all layers are updated during training. The input is the same as during pretraining, i.e., $60\mathrm{s}$ windows at $4~\mathrm{Hz}$ with three channels (tonic, phasic, and original EDA signal), with no resampling.
    We train with the Adam optimizer~\cite{kingma2014adam} and a class-weighted loss, to account for the class imbalance of some downstream tasks, consistently with the \texttt{balanced} class weighting used for the logistic regression in linear probing (\cref{tab:logistic_regression_hyperparams}). We train for a maximum of 50 epochs, with early stopping on the validation macro F1-score and a patience of 10 epochs. As for linear probing, we select the learning rate (grid: $5\!\times\!10^{-6}$, $2\!\times\!10^{-5}$, $6.8\!\times\!10^{-5}$, $1\!\times\!10^{-4}$) and the batch size (grid: 4, 8, 16) independently at each cross-validation iteration, using a 3-fold inner cross-validation with grid search.

    To fine-tune Mantis on the downstream tasks~\cite{Feofanov2025}, we use a learning rate of $2\!\times\!10^{-5}$, 100 epochs and the same early stopping criteria as we used for \modelname. For the rest of the hyperparameters, we follow the original authors' implementation, since these are the original authors' recommended settings.

    \subsection{Data Augmentations for EDA Data}
    We report in \cref{tab:eda_augmentations} the list of data augmentations used to train our \modelname foundation model with contrastive learning. We use the same set proposed by \citet{Matton2023}, which contains EDA-specific augmentations.

    \begin{table}[t]
    \centering
    \small
    \caption{Summary of all datasets in the EDAMAME collection with their corresponding data-sharing agreements. A \cmark\ indicates the data can be re-shared as part of the collection, a \xmark\ that it cannot.}
    \label{tab:appendix_datasets}

    \begin{tabularx}{\columnwidth}{lX}
        \toprule
        \textbf{Dataset Name} & \textbf{License} \\
        \midrule
        Amigos~\cite{miranda2018amigos} & \cmark - (CCBY 4) with original license \\
        APSync~\cite{Gashi2019} & \cmark - data sharing agreement \\
        BIG IDEAS~\cite{Bent2021} & \cmark - with original license (ODC-BY 1) \\
        BiHeartS~\cite{Abdalazim2025} & \cmark - data sharing agreement \\
        DREAMT~\cite{Wang} & \xmark - PhysioNet Restricted Health Data Use Agreement 1.5.0 \\
        Dynamics in the workplace~\cite{Lukan2018} & \cmark - data sharing agreement \\
        EDABE~\cite{llanes2023edabe} & \cmark - (CCBY 4) with original license \\
        EmoStress~\cite{gjoreski2017monitoring,gjoreski2016continuous} & \cmark - (CCBY 4) with original license \\
        EmpaticaE4Stress~\cite{Campanella2023} & \cmark - (CCBY 4) with original license \\
        EPM-E4~\cite{Garcia-Moreno2020} & \cmark - (CCBY 4) with original license \\
        HeartS~\cite{Abdalazim2023} & \cmark - data sharing agreement \\
        HHISS~\cite{xiao2025hhiss} & \xmark - no license specified \\
        LAUREATE~\cite{Laporte2023} & \cmark - data sharing agreement \\
        M2Sleep~\cite{Gashi2022a} & \cmark - data sharing agreement \\
        MEFAR~\cite{Derdiyok2024} & \cmark - (CCBY 4) with original license \\
        Multi-physio~\cite{lucas2026physiological} & \cmark - (CCBY 4) with original license \\
        Nurses' Stress~\cite{Hosseini2022} & \cmark - (ODbL 1) with original license \\
        PPG-Dalia~\cite{Reiss2019} & \cmark - (CCBY 4) with original license \\
        Regulation~\cite{azgomi2023regulation} & \cmark - (CCBY 4) with original license \\
        SEED~\cite{DiLascio2018} & \cmark - data sharing agreement \\
        SEED-II-Lab~\cite{DiLascio2018} & \cmark - data sharing agreement \\
        SEED-II-Wild~\cite{DiLascio2018} & \cmark - data sharing agreement \\
        Stress Predict~\cite{Iqbal2022} & \cmark - (MIT) with original license \\
        ToadStool~\cite{Svoren2020} & \cmark - (CCBY 4) with original license \\
        USILaughs~\cite{DiLascio2019} & \cmark - data sharing agreement \\
        WEEE~\cite{Gashi2022} & \cmark - (CCBY 4) with original license \\
        WESAD~\cite{Schmidt2018} & \cmark - (CCBY 4) with original license \\
        WESD~\cite{RafiulAmin2022} & \cmark - with original license (ODC-BY 1) \\
        Workplace~\cite{DiLascio2021} & \cmark - data sharing agreement \\
        \bottomrule
    \end{tabularx}
\end{table}

\begin{table}
    \centering
    \footnotesize
    \caption{Overview of the models and feature sets used in the experimental evaluation. We report the model size (number of trainable parameters) and the dimensionality of the feature space ($d$) used for linear probing. For the generalist foundation models, the reported size refers to the specific pretrained variant we use in all our experiments (\cref{app:handcrafted_baselines}).}
    \label{tab:model_sizes_dimensions}
    \begin{tabular}{lcc}
        \toprule
        \textbf{Model / Feature Set} & \textbf{Model Size} & \textbf{Feature Dim. ($d$)} \\
        \midrule
        \addlinespace[3pt]
        \rowcolor{gray!12}
        \multicolumn{3}{l}{\textbf{\scshape Handcrafted Methods}} \\
        \addlinespace[3pt]
        Generic HC & N/A & 12 \\
        EDA-specific HC & N/A & 45 \\

        \addlinespace[6pt]
        \rowcolor{gray!12}
        \multicolumn{3}{l}{\textbf{\scshape Generalist Foundation Models}} \\
        \addlinespace[3pt]
        Mantis~\cite{Feofanov2025} & $\sim$ 8 M & 768 \\
        Chronos~\cite{Ansari2024} (\emph{Small}) & $\sim$ 28 M & 1536 \\
        MOMENT~\cite{Goswami2024} (\emph{Large}) & $\sim$ 128 M & 1024 \\

        \addlinespace[6pt]
        \rowcolor{gray!12}
        \multicolumn{3}{l}{\textbf{\scshape Ours}} \\
        \addlinespace[3pt]
        \textbf{\modelname} & $\sim$ \textbf{1 M} & \textbf{64} \\
        \bottomrule
    \end{tabular}
\end{table}

\begin{table}
      \centering
      \footnotesize
      \caption{Hyperparameter Grid for Logistic Regression}
      \label{tab:logistic_regression_hyperparams}
      \begin{tabular}{ll}
        \toprule
        \textbf{Hyperparameter} & \textbf{Values} \\
        \midrule
        Regularization strength ($C$) & 0.01, 0.1, 1, 10 \\
        Solver & lbfgs, liblinear \\
        Penalty & L2 \\
        Max iterations & 10000 \\
        Class weight & balanced \\
        \bottomrule
      \end{tabular}
    \end{table}

    \begin{table}[t]
        \centering
        \scriptsize
        \caption{Mathematical formulations for the 15 base handcrafted features extracted from EDA time series signals ($x$) of length $N$. All features are computed independently for the tonic, phasic, and original EDA components. The ``Feature Set(s)'' column indicates whether the feature was used in the Generic baseline (12 total features) or the EDA-specific baseline (45 total features).}
        \label{tab:features_appendix}

        \renewcommand{\arraystretch}{1.8}
        \begin{tabular}{llc}
        \toprule
        \textbf{Feature} & \textbf{Mathematical Notation or Formula} & \textbf{Feature Set(s)} \\
        \midrule
        \addlinespace[6pt]
        \rowcolor{gray!12}
        \multicolumn{3}{l}{\textbf{\scshape Time-domain}} \\
        \addlinespace[3pt]
        Mean & $\frac{1}{N}\sum_{i=1}^{N} x_i$ & Generic \& EDA-specific \\
        Minimum & $\min(x_1, x_2, \ldots, x_N)$ & Generic \& EDA-specific \\
        Maximum & $\max(x_1, x_2, \ldots, x_N)$ & Generic \& EDA-specific \\
        Standard Deviation & $\sqrt{\frac{1}{N-1}\sum_{i=1}^{N} (x_i - \bar{x})^2}$ & Generic \& EDA-specific \\
        \midrule
        Dynamic Range & $\max(x_1, x_2, \ldots, x_N) - \min(x_1, x_2, \ldots, x_N)$ & EDA-specific only \\
        Slope & $\frac{x_N - x_1}{N-1}$ & EDA-specific only \\
        Absolute Value of Slope & $\left|\frac{x_N - x_1}{N-1}\right|$ & EDA-specific only \\
        Mean of the First Derivative & $\frac{1}{N-1}\sum_{i=1}^{N-1} (x_{i+1} - x_i)$ & EDA-specific only \\
        Std. Dev. of the First Derivative & $\sqrt{\frac{1}{N-2}\sum_{i=1}^{N-1} ((x_{i+1} - x_i) - \bar{x'})^2}$ & EDA-specific only \\
        Number of EDA Peaks & Count of local maxima in the window & EDA-specific only \\
        Amplitude of EDA Peaks & Mean amplitude of local maxima in the window & EDA-specific only \\
        \midrule
        \addlinespace[6pt]
        \rowcolor{gray!12}
        \multicolumn{3}{l}{\textbf{\scshape Frequency Domain (Fast Fourier Transform)}} \\
        \addlinespace[3pt]
        Direct Current (DC) & $X_0$ & EDA-specific only \\
        Sum of Frequency Coefficients & $\sum_{k=1}^{N} |X_k|$ & EDA-specific only \\
        Information Entropy & $-\sum_{k=1}^{N} P(X_k) \log_2(P(X_k))$ & EDA-specific only \\
        Spectral Energy & $\sum_{k=1}^{N} |X_k|^2$ & EDA-specific only \\
        \bottomrule
        \end{tabular}
    \end{table}

    \begin{table}[t]
      \centering
      \footnotesize
      \caption{Summary of Electrodermal Activity (EDA) Data Augmentations used to train the \modelname foundation model. The table details both EDA-specific transforms designed to isolate physiological components or simulate artifacts, and generic time series transforms. The set is the same as that proposed by \citet{Matton2023}.}
      \label{tab:eda_augmentations}
      \renewcommand{\arraystretch}{1.2}
      \begin{tabularx}{\textwidth}{l l X l}
        \toprule
        \textbf{Augmentation} & \textbf{Type} & \textbf{Description} & \textbf{Parameter Range} \\
        \midrule

        \addlinespace[6pt]
        \rowcolor{gray!12}
        \multicolumn{4}{l}{\textbf{\scshape Frequency Domain \& Component Isolation}} \\
        \addlinespace[3pt]
        Low-Pass Filter & EDA-Specific & Extracts tonic component; removes high-freq noise. & Cutoff $f \in [0.25, 1.0]$ Hz  \\
        High-Pass Filter & EDA-Specific & Extracts phasic component; removes slow drifts. & Cutoff $f \in [0.05, 0.25]$ Hz  \\
        Band-Pass Filter & EDA-Specific & Isolates information-rich EDA frequency bands. & Pass $f \in [0.05, 0.25]$ Hz  \\
        Band-Stop Filter & EDA-Specific & Rejects specific frequency bands. & Reject $f \in [0.75, 1.0]$ Hz  \\
        High Freq. Noise & EDA-Specific & Adds Gaussian noise only to frequencies $>1$Hz. & Noise $\sigma \in [0, 0.5]$  \\

        \addlinespace[6pt]
        \rowcolor{gray!12}
        \multicolumn{4}{l}{\textbf{\scshape Artifact Simulation}} \\
        \addlinespace[3pt]
        Jump Artifact & EDA-Specific & Simulates abrupt sensor movement/displacement. & Jump $\in [0.01, 0.2] \mu S$  \\
        Loose Sensor & EDA-Specific & Simulates electrode contact loss (signal drop). & Duration $t \in [5, 20]$ s  \\

        \addlinespace[6pt]
        \rowcolor{gray!12}
        \multicolumn{4}{l}{\textbf{\scshape Thermoregulation Simulation}} \\
        \addlinespace[3pt]
        Tonic Const. Scale & EDA-Specific & Scales tonic component (simulates constant temp). & Factor $s \in [0.25, 2]$  \\
        Tonic Amp. Warp & EDA-Specific & Time-varying scale of tonic component (changing conditions). & Spline $\sigma \in [0.01, 0.05]$  \\

        \addlinespace[6pt]
        \rowcolor{gray!12}
        \multicolumn{4}{l}{\textbf{\scshape Generic Time Series}} \\
        \addlinespace[3pt]
        Amp. Const. Scale & Generic & Applies constant scaling factor to the raw signal. & Factor $s \in [0.25, 2]$  \\
        Amplitude Warp & Generic & Applies smooth, time-varying scaling to raw signal. & Spline $\sigma \in [0.01, 0.05]$\\
        Gaussian Noise & Generic & Adds random Gaussian noise to the raw signal. & $\sigma \in [0, 0.5]$  \\
        Time Shift & Generic & Shifts the signal window forward or backward. & Shift $t \in [5, 45]$ s  \\
        Temporal Cutout & Generic & Masks/zeroes out a random sub-window of the signal. & Cutout $t \in [5, 15]$ s \\
        Time Warp & Generic & Perturbs temporal dimension (local stretch/compress). & Spline $\sigma \in [0.01, 0.1]$ \\
        Permutation & Generic & Slices signal and randomly reorders sub-windows. & Segments $n \in [2, 6]$ \\
        Signal Flip & Generic & Inverts the signal over its amplitude dimension. & N/A \\

        \bottomrule
      \end{tabularx}
    \end{table}

\clearpage
\section{Sensitivity \& Additional Model Studies}
\label{app:ablation}

We perform sensitivity and additional studies to find the configuration of our \modelname foundation model. In this section, we report details for: implementation of an additional neural network architecture, i.e., a Masked Autoencoder (MAE); studies on model size and hyperparameters for the EfficientNet architecture chosen for the \modelname foundation model, trained using contrastive learning.

\subsection{Experiments with Masked Autoencoders (MAEs)}
We train a reconstruction-based masked autoencoder (MAE), based on a vision transformer (ViT)~\cite{Dosovitskiy2020}. \citet{Narayanswamy2024} also used a masked autoencoder to train a foundation model for physiological data.
We train the model using the following loss: let $\mathbf{x} \in \mathbb{R}^{T \times C}$ denote the input EDA signal, partitioned into $N$ non-overlapping patches of size $P$, and let $\hat{\mathbf{x}}_i$ and $\mathbf{x}_i$ denote the reconstructed and original $i$-th patch, respectively.
Given a binary mask $\mathbf{m} \in \{0,1\}^N$, where $m_i = 1$ indicates a masked patch and $m_i = 0$ a visible one, the reconstruction loss is defined as
\begin{equation}
\mathcal{L}
= \alpha \, \mathcal{L}_{\text{masked}} + (1-\alpha)\, \mathcal{L}_{\text{visible}},
\end{equation}
with
\begin{align}
\mathcal{L}_{\text{masked}} &= \frac{1}{\sum_i m_i} \sum_{i=1}^N m_i \, \ell(\hat{\mathbf{x}}_i, \mathbf{x}_i), \\
\mathcal{L}_{\text{visible}} &= \frac{1}{\sum_i (1-m_i)} \sum_{i=1}^N (1-m_i)\, \ell(\hat{\mathbf{x}}_i, \mathbf{x}_i),
\end{align}
where $\ell(\cdot,\cdot)$ denotes the mean absolute error (not to be confused with the masked autoencoder, MAE) computed within each patch, and $\alpha \in [0,1]$ controls the relative importance of masked versus visible patch reconstruction.

\begin{figure}[t]
    \centering
    \includegraphics[width=0.5\linewidth]{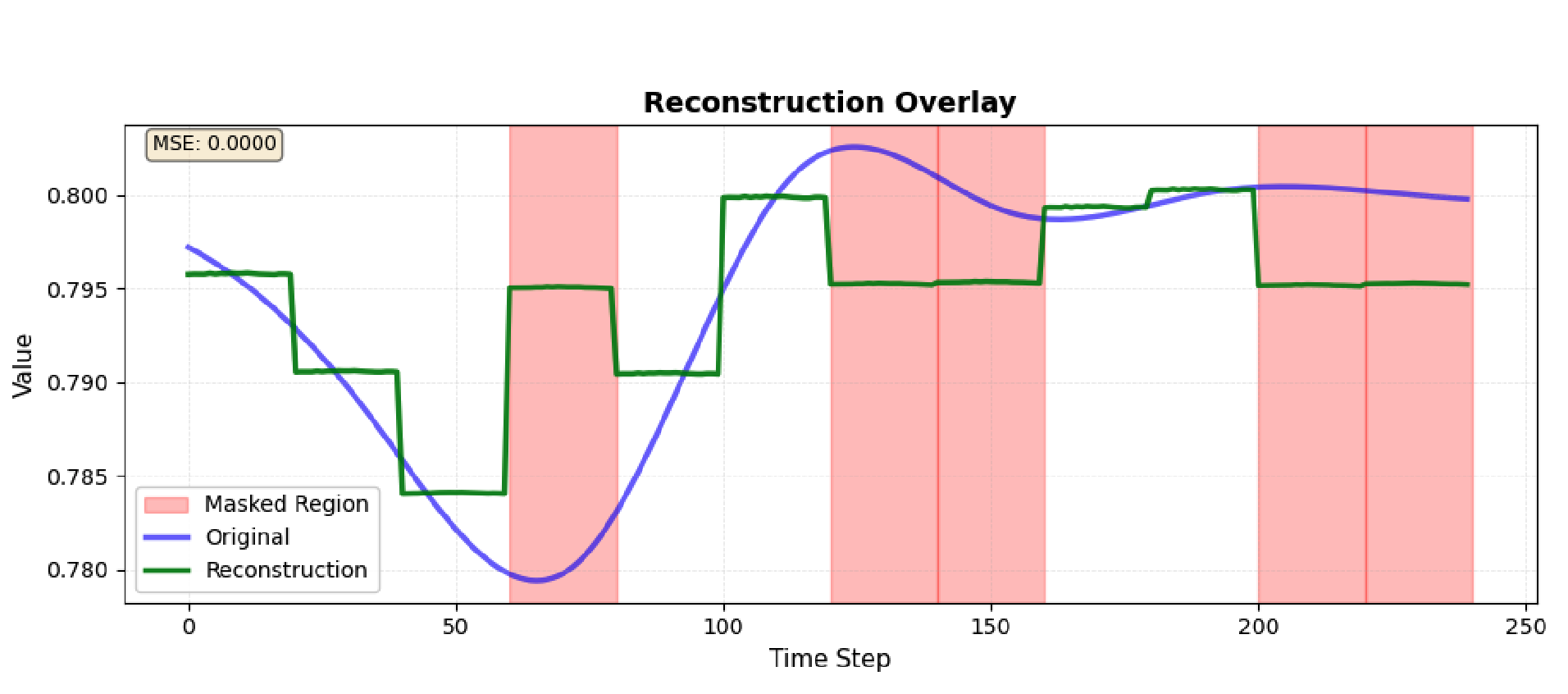}
    \caption{Example of EDA signal reconstruction using MAE.}
    \Description{A line plot of a 60-second electrodermal activity window, showing the original signal overlaid with the signal reconstructed by the masked autoencoder.}

    \label{fig:example-reconstruction}
\end{figure}

\begin{figure}[t]
    \centering
    \includegraphics[width=0.8\linewidth]{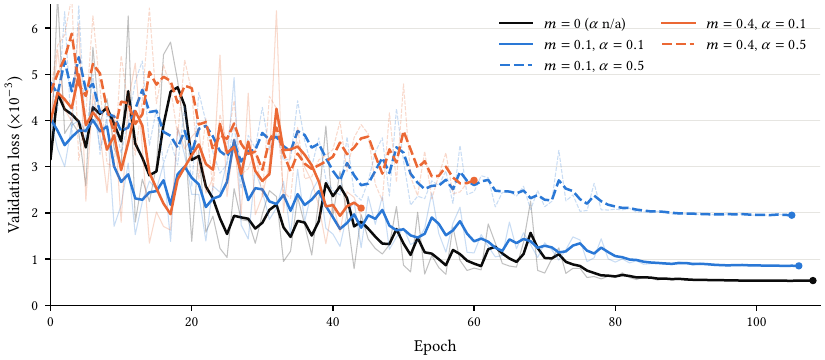}
    \caption{Validation loss during training for different configurations of masking and masking weight.}
    \Description{Line plots of validation loss against training epoch, with one curve per combination of masking ratio and masking weight.}

    \label{fig:mae_training_curve}
\end{figure}

We experiment with the following configurations: $m \in \{0, 0.1, 0.4 \}$ and $\alpha \in \{0.1, 0.5\}$.
In \cref{fig:mae_training_curve} we report the validation loss curve during training for the configurations tested, and in \cref{fig:example-reconstruction} we show an example of a signal reconstruction at train end, over a validation sample.
We also perform downstream evaluation on the BiHeartS dataset: however, no configuration leads to performance, in terms of balanced accuracy, above that of the dummy classifier.

Given these findings, we believe that masked autoencoders are not optimal for the specific EDA data present in the \dataname collection of datasets.

\clearpage

\subsection{Sensitivity Study on Model Size}
\label{app:model_size}

We adopt an EfficientNet architecture for our \modelname foundation model, which we train using contrastive learning on EDA data. We perform a sensitivity study on the model size. To this end, we train four versions of the \modelname foundation model: tiny ($\sim 50k$ parameters), small ($\sim 200k$ parameters), base ($\sim 1M$ parameters) and large ($\sim 2M$ parameters). We report the details of their individual hyperparameters in \cref{tab:efficientnet_hyperparameters}. We pretrain all of these models using the complete pretrain part of the \dataname collection, as in the final model described in \cref{sec:model_arch}; we also use the same pretrain configurations, e.g., number of epochs, as the final model, as reported in \cref{sec:training_info}.

In order to avoid bias in the result presentation, we compare the performance of these different configurations using only one downstream task: cognitive load/relaxation from USILaughs, which is out-of-distribution from the pretrain data and a task relevant for EDA.

We report the results in \cref{tab:model_size}. We notice that the larger the model, the higher the balanced accuracy. We also notice that there is no difference between the \emph{base} and \emph{large} models, suggesting that increasing beyond the base model size, with the specific pretrain data that we have, does not lead to an increase in performance.

    \begin{table}[t]
    \centering
    \caption{Hyperparameter configurations for the four EfficientNet model variants. Across all versions, several hyperparameters remain fixed, including the number of input channels ($C_{in} = 3$), the input sequence/segment length ($L = 240$), the base convolution stride ($S = 2$), and the use of the Swish activation function.}
    \label{tab:efficientnet_hyperparameters}
    \begin{tabular}{l c c c c}
    \toprule
    \textbf{Hyperparameter} & \textbf{Tiny} & \textbf{Small} & \textbf{Base} & \textbf{Large} \\
    \midrule
    Stem Channels & $32$ & $48$ & $64$ & $96$ \\
    MBConv Channels & $32$ & $48$ & $64$ & $96$ \\
    MBConv Blocks & $4$ & $8$ & $16$ & $24$ \\
    Head Channels & $96$ & $160$ & $248$ & $384$ \\
    Embedding Dim & $32$ & $48$ & $64$ & $96$ \\
    Kernel Size & $5$ & $7$ & $9$ & $11$ \\
    Dropout & $0.1$ & $0.1$ & $0.1$ & $0.1$ \\
    \bottomrule
    \end{tabular}
\end{table}

        \begin{table}[t]
        \centering
        \caption{Sensitivity analysis of \modelname model size under leave-one-participant-out (LOPO) cross-validation.}
        \label{tab:model_size}
        \renewcommand{\arraystretch}{1.2}
        \begin{tabular}{ll | c| cc| ccccc}
        \toprule
        \textbf{Dataset} & \textbf{Task}
        & \textbf{Random}
        & \textbf{Gen.}
        & \textbf{EDA}
        & \textbf{Tiny}
        & \textbf{Small}
        & \textbf{Base}
        & \textbf{Large} \\
        \midrule

        USILaughs & Cog. load / relax
        & $.50_{.00}$ & $.66_{.03}$ & $.70_{.04}$
        & $.61_{.03}$ & $.64_{.03}$ & $.71_{.05}$ & $.71_{.05}$
         \\

        \bottomrule
        \end{tabular}

\end{table}

    \subsection{Sensitivity Study on Dataset Size}\label{app:dataset_size}

    In what is referred to as \emph{foundation model scaling laws}, researchers have shown that the size of the pretrain dataset plays a significant role in the downstream performance of foundation models~\cite{Narayanswamy2024,shuai2026osf}. To this end, we perform a sensitivity study. In particular, we consider the base \modelname model and pretrain it on different subsets of the \dataname-train subcollection: 5\%, 20\%, 50\% and 75\%.

    Similarly to the study in \cref{app:model_size}, we compare the performance of the different configurations using a single downstream task, cognitive load/relaxation on the USILaughs dataset. We use the same pretrain setup.

    We report the results in \cref{tab:data_size}, in terms of balanced accuracy. We notice that our foundation model follows the foundation model scaling laws, as does similar work in the literature, e.g., \cite{Narayanswamy2024,shuai2026osf}.

    From these results we also conclude that the size of the pretrain dataset, \dataname, is indeed necessary to achieve results higher than those of existing approaches.

    \begin{table}[t]
            \centering
            \caption{Results for \emph{sensitivity analysis of dataset size on \modelname training}. Experiments were run using the base \modelname model. Random undersampling of the complete pretrain \dataname dataset was used. Reported is the \emph{balanced accuracy}.}
            \label{tab:data_size}
            \renewcommand{\arraystretch}{1.3}
            \begin{tabularx}{\textwidth}{lX | c | cc | ccccc}
            \toprule
            \textbf{Dataset} & \textbf{Binary Task}
            & \textbf{Random}
            & \textbf{Gen.}
            & \textbf{EDA-spec.}
            & \textbf{5\%}
            & \textbf{20\%}
            & \textbf{50\%}
            & \textbf{75\%}
            & \textbf{\modelname (100\%)}\\
            \midrule
            USILaughs & Cog. load/relax
            & $.50_{.00} $ & $.66_{.03}$ & $.70_{.04}$  & $.61_{.03}$ & $ .63_{.05}$ & $ .64_{.03} $ & $.68_{.04}$ & $.71_{.05}$ \\
            \bottomrule
            \end{tabularx}
        \end{table}

    \subsection{Sensitivity Study on Training Loss Pairs}\label{app:user-wise pairs}

        We train \modelname using as loss function the InfoNCE contrastive loss, where positive pairs are defined as two different augmentations of the input and the negative pairs are all other possible comparisons inside the batch.
        However, researchers working on foundation models for wearable physiological data have also used different ways to generate the positive pairs. For example, \citet{Abbaspourazad2024}, who used an EfficientNet and InfoNCE loss as well, created positive pairs from signals belonging to the same user, and negative pairs as signals belonging to different users.
        They use this approach for foundation models of PPG and ECG data.

        We pretrain our model using user-pairs, as did \citet{Abbaspourazad2024}. The pretraining setup is the same as the base \modelname model, using 100\% of the pretrain data. Similarly to the other sensitivity studies, we compare the performance on the USILaughs cognitive load/relaxation task only, to avoid selection bias.
        We report the results in \cref{tab:user-pair}. We notice that the performance is lower than that of the sample-pair method, which we use for \modelname. Importantly, we notice that the model overfitted after only 5 training epochs: we speculate that this is due to the inherent variability in EDA data. While PPG and ECG signals have distinct user patterns, which for example allowed researchers to use these signals for user identification~\cite{Hinatsu2020PhotoplethysmographicSubjectIdentification}, the same is not true for EDA. Indeed, we believe that the model manages to understand, to a limited degree, the underlying time series elements of EDA signals, but cannot differentiate well between signals from different users.

        \begin{table}[t]
            \centering
            \caption{Results for \emph{sensitivity analysis of training loss pairs on \modelname}. Experiments were run using the base \modelname model trained with user-assignment pairs on the complete pretrain \dataname dataset. Reported is the \emph{balanced accuracy}.}
            \label{tab:user-pair}
            \renewcommand{\arraystretch}{1.3}
            \begin{tabularx}{\textwidth}{lX | c | cc | cc}
            \toprule
            \textbf{Dataset} & \textbf{Binary Task}
            & \textbf{Random}
            & \textbf{Gen.}
            & \textbf{EDA-spec.}
            & \textbf{User-pair}
            & \textbf{Sample-pair (paper)} \\
            \midrule

            USILaughs & Cog. load/relax
            & $.50_{.00} $ & $.66_{.03}$ & $.70_{.04}$  & $.68_{.03}$ & $.71_{.05}$ \\
            \bottomrule
            \end{tabularx}
        \end{table}

    \subsection{Fine-Tuning Mantis on the Pretrain Dataset}\label{app:mantis-finetune}

    We fine-tune the Mantis foundation model~\cite{Feofanov2025} on our pretrain data. We perform this experiment on Mantis only since it is the generalist foundation model that achieves the highest performance in our experimental evaluation. We fine-tune using the same pretrain setup as used by the Mantis authors\footnote{\url{https://github.com/vfeofanov/mantis}} and using the complete \dataname-train pretrain dataset.

    Similarly to the other experiments in this Appendix, we report the results in terms of balanced accuracy on the USILaughs cognitive load/relaxation task. We report the results in \cref{tab:mantis}. We notice that the results of the fine-tuned model are lower than those of the generalist Mantis and our \modelname model. We speculate that this is due to catastrophic forgetting when fine-tuning on a dataset significantly different from the pretrain data~\cite{li2024revisitingcatastrophicforgettinglarge}.

    \begin{table}[t]
            \centering
            \small
            \caption{Results for \textbf{analysis of Mantis fine tune}. Experiments were run using the complete Mantis model, as downloaded from HuggingFace, and fine-tuned on our pretrain data.}
            \label{tab:mantis}
            \renewcommand{\arraystretch}{1.3}
            \begin{tabularx}{\textwidth}{lX | c | cc | ccc}
            \toprule
            \textbf{Dataset} & \textbf{Binary Task}
            & \textbf{Random}
            & \textbf{Gen.}
            & \textbf{EDA-spec.}
            & \textbf{Mantis fine tune}
            & \textbf{Mantis (no fine tune)}
            & \textbf{\modelname (paper)}\\
            \midrule

            USILaughs & Cog. load/relax
            & $.50_{.00} $ & $.66_{.03}$ & $.70_{.04}$  & $ .68_{.05}$ & $ .72_{.05}$ & $.71_{.05}$ \\

            \bottomrule
            \end{tabularx}
        \end{table}

\section{Additional Results}
\label{app:additional_results}

\paragraph{Computational complexity analysis}
We report in \cref{tab:extractor-performance} the detailed performance metrics for the computational complexity analysis. This table mirrors the results in \cref{fig:computational_complexity}.

\begin{table}[t]
\centering
\caption{CPU Performance Comparison of Extractors. SE denotes the standard error.}
\label{tab:extractor-performance}
\small
\begin{tabular}{lrr}
\toprule
\textbf{Extractor} & \textbf{CPU Time (mean) [ms]} & \textbf{CPU Time (SE) [ms]} \\
\midrule
MOMENT (128M)           & 2223.53 & 161.11 \\
Chronos (28M)           & 721.34  & 28.91  \\
Mantis (8M)             & 350.23  & 39.06  \\
\modelname (1M) & 184.56  & 3.12   \\
EDA-specific hc         & 35.39   & 1.40   \\
generic hc              & 0.09    & 0.01   \\
\bottomrule
\end{tabular}
\end{table}

\paragraph{Complete results for the downstream evaluation}
We report in \cref{tab:binary_results_lopo} and \cref{tab:binary_results_tacv} the complete results for our experimental evaluation. Specifically, we report the results in terms of balanced accuracy, MCC~\cite{Chicco2020} and F1-score. We report the results for both validation methods used, i.e., LOPO and TA cross-validations.
For the TA cross-validation, we do not report results for \emph{cognitive-load/relaxation} on USILaughs~\cite{DiLascio2019}, for \emph{relax/video} on Multi-physio~\cite{lucas2026physiological}, or for the two Amigos~\cite{miranda2018amigos} tasks: in all four cases the amount of data available per user is insufficient to perform the chronological $2/3$--$1/3$ split that TA cross-validation requires.

\paragraph{Friedman test ranking}
We report in \cref{tab:friedman-ranks} the full model ranking, as computed using the Friedman test. The rankings are shown with 95\% confidence intervals (bootstrap with $2 \times 10^6$ iterations). We notice that the Mantis and \modelname confidence intervals are nearly identical.

\begingroup
\tiny
\setlength{\tabcolsep}{3pt}
\renewcommand{\arraystretch}{0.9}
\begin{longtable}{ll l ccc}
\caption{Results of binary classification experiments, using the Leave-One-Participant-Out (LOPO) cross-validation. Results shown as \texttt{metric $\pm$ standard error}.}
\label{tab:binary_results_lopo} \\
\toprule
\multirow{2}{*}{\textbf{Dataset}} &
\multirow{2}{*}{\textbf{Binary Task}} &
\multirow{2}{*}{\textbf{Model}} &
\multicolumn{3}{c}{\textbf{Metrics}} \\
\cmidrule(lr){4-6}
 & & & \textbf{Balanced Acc.} & \textbf{MCC} & \textbf{F1} \\
\midrule
\endfirsthead
\toprule
\textbf{Dataset} & \textbf{Binary Task} & \textbf{Model} &
\textbf{Balanced Acc.} & \textbf{MCC} & \textbf{F1} \\
\midrule
\endhead
\midrule
\multicolumn{6}{r}{\emph{continued on next page}} \\
\endfoot
\bottomrule
\endlastfoot

\multirow{7}{*}{{DREAMT}} &
\multirow{7}{*}{{Deep Sleep/REM}}
  & Dummy         & $0.73 \pm 0.05$ & $0.00 \pm 0.01$ & $0.20 \pm 0.05$ \\
 & & Gen. HC        & $0.64 \pm 0.07$ & $0.01 \pm 0.02$ & $0.12 \pm 0.02$ \\
 & & EDA-spec. HC  & $0.54 \pm 0.02$ & $0.00 \pm 0.01$ & $0.13 \pm 0.02$ \\
 & & Mantis        & $0.58 \pm 0.03$ & $0.02 \pm 0.02$ & $0.13 \pm 0.02$ \\
 & & MOMENT        & $0.64 \pm 0.02$ & $0.06 \pm 0.02$ & $0.18 \pm 0.03$ \\
 & & Chronos       & $0.63 \pm 0.02$ & $0.03 \pm 0.02$ & $0.17 \pm 0.03$ \\
 & & \modelname    & $0.56 \pm 0.03$ & $0.04 \pm 0.02$ & $0.16 \pm 0.03$ \\
\midrule

\multirow{9}{*}{{DREAMT}} &
\multirow{9}{*}{{Sleep/Wake}}
  & Dummy         & $0.48 \pm 0.00$ & $-0.04 \pm 0.00$ & $0.51 \pm 0.01$ \\
 & & Gen. HC        & $0.70 \pm 0.01$ & $0.40 \pm 0.03$ & $0.68 \pm 0.02$ \\
 & & EDA-spec. HC  & $0.74 \pm 0.01$ & $0.48 \pm 0.02$ & $0.72 \pm 0.02$ \\
 & & Mantis        & $0.76 \pm 0.01$ & $0.52 \pm 0.02$ & $0.74 \pm 0.02$ \\
 & & MOMENT        & $0.73 \pm 0.01$ & $0.44 \pm 0.02$ & $0.71 \pm 0.01$ \\
 & & Chronos       & $0.78 \pm 0.01$ & $0.55 \pm 0.02$ & $0.75 \pm 0.02$ \\
 & & \modelname    & $0.75 \pm 0.01$ & $0.50 \pm 0.02$ & $0.73 \pm 0.02$ \\
 & & \modelname\ (FT) & $0.76 \pm 0.01$ & $0.53 \pm 0.02$ & $0.79 \pm 0.01$ \\
 & & Mantis (FT)   & $0.75 \pm 0.01$ & $0.52 \pm 0.02$ & $0.77 \pm 0.02$ \\
\midrule

\multirow{9}{*}{{HHISS}} &
\multirow{9}{*}{{Stress/calm}}
  & Dummy         & $0.50 \pm 0.00$ & $-0.00 \pm 0.00$ & $0.24 \pm 0.03$ \\
 & & Gen. HC        & $0.56 \pm 0.02$ & $0.12 \pm 0.04$ & $0.44 \pm 0.03$ \\
 & & EDA-spec. HC  & $0.64 \pm 0.01$ & $0.28 \pm 0.03$ & $0.55 \pm 0.02$ \\
 & & Mantis        & $0.63 \pm 0.02$ & $0.26 \pm 0.03$ & $0.53 \pm 0.02$ \\
 & & MOMENT        & $0.55 \pm 0.01$ & $0.09 \pm 0.02$ & $0.44 \pm 0.02$ \\
 & & Chronos       & $0.59 \pm 0.01$ & $0.17 \pm 0.03$ & $0.48 \pm 0.02$ \\
 & & \modelname    & $0.60 \pm 0.02$ & $0.21 \pm 0.03$ & $0.53 \pm 0.02$ \\
 & & \modelname\ (FT) & $0.60 \pm 0.02$ & $0.20 \pm 0.02$ & $0.51 \pm 0.02$ \\
 & & Mantis (FT)   & $0.56 \pm 0.01$ & $0.11 \pm 0.01$ & $0.42 \pm 0.01$ \\
\midrule

\multirow{9}{*}{{HeartS}} &
\multirow{9}{*}{{Sleep/Wake}}
  & Dummy         & $0.50 \pm 0.00$ & $0.00 \pm 0.00$ & $0.08 \pm 0.02$ \\
 & & Gen. HC        & $0.66 \pm 0.04$ & $0.28 \pm 0.08$ & $0.32 \pm 0.06$ \\
 & & EDA-spec. HC  & $0.70 \pm 0.03$ & $0.32 \pm 0.07$ & $0.36 \pm 0.06$ \\
 & & Mantis        & $0.74 \pm 0.03$ & $0.37 \pm 0.07$ & $0.41 \pm 0.06$ \\
 & & MOMENT        & $0.70 \pm 0.02$ & $0.27 \pm 0.03$ & $0.33 \pm 0.04$ \\
 & & Chronos       & $0.73 \pm 0.03$ & $0.34 \pm 0.05$ & $0.39 \pm 0.06$ \\
 & & \modelname    & $0.72 \pm 0.03$ & $0.30 \pm 0.05$ & $0.35 \pm 0.05$ \\
 & & \modelname\ (FT) & $0.71 \pm 0.04$ & $0.40 \pm 0.04$ & $0.41 \pm 0.04$ \\
 & & Mantis (FT)   & $0.71 \pm 0.05$ & $0.47 \pm 0.04$ & $0.45 \pm 0.03$ \\
\midrule

\multirow{9}{*}{{WESAD}} &
\multirow{9}{*}{{Low/High Arousal}}
  & Dummy         & $0.58 \pm 0.05$ & $-0.02 \pm 0.03$ & $0.05 \pm 0.03$ \\
 & & Gen. HC        & $0.61 \pm 0.04$ & $0.10 \pm 0.07$ & $0.26 \pm 0.07$ \\
 & & EDA-spec. HC  & $0.63 \pm 0.04$ & $0.13 \pm 0.06$ & $0.33 \pm 0.07$ \\
 & & Mantis        & $0.56 \pm 0.04$ & $0.06 \pm 0.07$ & $0.27 \pm 0.07$ \\
 & & MOMENT        & $0.66 \pm 0.04$ & $0.17 \pm 0.10$ & $0.38 \pm 0.07$ \\
 & & Chronos       & $0.66 \pm 0.05$ & $0.17 \pm 0.10$ & $0.38 \pm 0.07$ \\
 & & \modelname    & $0.61 \pm 0.05$ & $0.11 \pm 0.06$ & $0.27 \pm 0.06$ \\
 & & \modelname\ (FT) & $0.70 \pm 0.04$ & $0.23 \pm 0.08$ & $0.37 \pm 0.08$ \\
 & & Mantis (FT) & $0.61 \pm 0.05$ & $0.12 \pm 0.05$ & $0.28 \pm 0.07$ \\
\midrule

\multirow{8}{*}{{WESAD}} &
\multirow{8}{*}{{Low/High Valence}}
  & Dummy         & $0.51 \pm 0.06$ & $0.02 \pm 0.05$ & $0.55 \pm 0.05$ \\
 & & Gen. HC        & $0.55 \pm 0.15$ & $0.06 \pm 0.05$ & $0.60 \pm 0.06$ \\
 & & EDA-spec. HC  & $0.57 \pm 0.06$ & $0.10 \pm 0.07$ & $0.63 \pm 0.07$ \\
 & & Mantis        & $0.56 \pm 0.06$ & $0.07 \pm 0.08$ & $0.59 \pm 0.08$ \\
 & & MOMENT        & $0.63 \pm 0.06$ & $0.00 \pm 0.05$ & $0.83 \pm 0.05$ \\
 & & Chronos       & $0.56 \pm 0.04$ & $0.05 \pm 0.05$ & $0.60 \pm 0.07$ \\
 & & \modelname    & $0.59 \pm 0.03$ & $0.08 \pm 0.05$ & $0.64 \pm 0.06$ \\
\midrule
\multirow{7}{*}{{APSync}} &
\multirow{7}{*}{{Low/High engagement}}
  & Dummy         & $0.46 \pm 0.03$ & $-0.09 \pm 0.06$ & $0.24 \pm 0.07$ \\
 & & Gen. HC        & $0.61 \pm 0.13$ & $0.16 \pm 0.14$ & $0.60 \pm 0.10$ \\
 & & EDA-spec. HC  & $0.63 \pm 0.16$ & $0.25 \pm 0.21$ & $0.70 \pm 0.12$ \\
 & & Mantis        & $0.62 \pm 0.11$ & $0.27 \pm 0.25$ & $0.71 \pm 0.10$ \\
 & & MOMENT        & $0.60 \pm 0.09$ & $0.24 \pm 0.20$ & $0.65 \pm 0.10$ \\
 & & Chronos       & $0.48 \pm 0.16$ & $0.02 \pm 0.05$ & $0.50 \pm 0.12$ \\
 & & \modelname    & $0.55 \pm 0.07$ & $0.10 \pm 0.15$ & $0.42 \pm 0.11$ \\
\midrule

\multirow{9}{*}{{USILaughs}} &
\multirow{9}{*}{{Cog. load/relax}}
  & Dummy         & $0.50 \pm 0.00$ & $0.00 \pm 0.00$ & $0.80 \pm 0.00$ \\
 & & Gen. HC        & $0.66 \pm 0.03$ & $0.32 \pm 0.07$ & $0.72 \pm 0.02$ \\
 & & EDA-spec. HC  & $0.70 \pm 0.04$ & $0.40 \pm 0.09$ & $0.70 \pm 0.06$ \\
 & & Mantis        & $0.72 \pm 0.05$ & $0.43 \pm 0.10$ & $0.76 \pm 0.05$ \\
 & & MOMENT        & $0.60 \pm 0.05$ & $0.20 \pm 0.09$ & $0.61 \pm 0.06$ \\
 & & Chronos       & $0.67 \pm 0.03$ & $0.35 \pm 0.07$ & $0.57 \pm 0.06$ \\
 & & \modelname    & $0.71 \pm 0.05$ & $0.42 \pm 0.11$ & $0.76 \pm 0.05$ \\
 & & \modelname\ (FT) & $0.74 \pm 0.04$ & $0.48 \pm 0.09$ & $0.75 \pm 0.05$ \\
 & & Mantis (FT)   & $0.72 \pm 0.04$ & $0.43 \pm 0.09$ & $0.83 \pm 0.03$ \\
\midrule

\multirow{9}{*}{{EDABE}} &
\multirow{9}{*}{{Artifact}}
  & Dummy            & $0.50 \pm 0.00$ & $-0.00 \pm 0.01$ & $0.58 \pm 0.04$ \\
 & & Gen. HC           & $0.64 \pm 0.01$ & $0.30 \pm 0.01$ & $0.58 \pm 0.03$ \\
 & & EDA-spec. HC     & $0.68 \pm 0.01$ & $0.33 \pm 0.02$ & $0.57 \pm 0.03$ \\
 & & Mantis           & $0.72 \pm 0.01$ & $0.38 \pm 0.02$ & $0.66 \pm 0.03$ \\
 & & MOMENT           & $0.70 \pm 0.01$ & $0.33 \pm 0.02$ & $0.63 \pm 0.03$ \\
 & & Chronos          & $0.75 \pm 0.02$ & $0.50 \pm 0.04$ & $0.75 \pm 0.04$ \\
 & & \modelname       & $0.69 \pm 0.01$ & $0.33 \pm 0.02$ & $0.61 \pm 0.03$ \\
 & & \modelname\ (FT) & $0.75 \pm 0.01$ & $0.45 \pm 0.02$ & $0.69 \pm 0.03$ \\
 & & Mantis (FT)      & $0.72 \pm 0.01$ & $0.39 \pm 0.03$ & $0.69 \pm 0.03$ \\
\midrule

\multirow{9}{*}{{Multi-physio}} &
\multirow{9}{*}{{Relax/Video}}
  & Dummy            & $0.49 \pm 0.02$ & $-0.01 \pm 0.05$ & $0.35 \pm 0.01$ \\
 & & Gen. HC           & $0.74 \pm 0.07$ & $0.49 \pm 0.16$ & $0.63 \pm 0.14$ \\
 & & EDA-spec. HC     & $0.79 \pm 0.07$ & $0.61 \pm 0.15$ & $0.70 \pm 0.13$ \\
 & & Mantis           & $0.74 \pm 0.06$ & $0.54 \pm 0.13$ & $0.78 \pm 0.07$ \\
 & & MOMENT           & $0.69 \pm 0.06$ & $0.40 \pm 0.13$ & $0.72 \pm 0.07$ \\
 & & Chronos          & $0.56 \pm 0.06$ & $0.13 \pm 0.11$ & $0.60 \pm 0.05$ \\
 & & \modelname       & $0.75 \pm 0.06$ & $0.53 \pm 0.13$ & $0.73 \pm 0.10$ \\
 & & \modelname\ (FT) & $0.75 \pm 0.06$ & $0.52 \pm 0.14$ & $0.55 \pm 0.14$ \\
 & & Mantis (FT)      & $0.81 \pm 0.06$ & $0.66 \pm 0.12$ & $0.87 \pm 0.04$ \\
\midrule

\multirow{9}{*}{{Amigos}} &
\multirow{9}{*}{{Arousal High/Low}}
  & Dummy            & $0.50 \pm 0.00$ & $0.04 \pm 0.03$ & $0.32 \pm 0.07$ \\
 & & Gen. HC           & $0.46 \pm 0.03$ & $-0.01 \pm 0.01$ & $0.54 \pm 0.08$ \\
 & & EDA-spec. HC     & $0.48 \pm 0.03$ & $0.01 \pm 0.07$ & $0.41 \pm 0.07$ \\
 & & Mantis           & $0.55 \pm 0.04$ & $0.13 \pm 0.06$ & $0.43 \pm 0.08$ \\
 & & MOMENT           & $0.49 \pm 0.01$ & $0.01 \pm 0.01$ & $0.02 \pm 0.02$ \\
 & & Chronos          & $0.50 \pm 0.03$ & $0.05 \pm 0.06$ & $0.32 \pm 0.08$ \\
 & & \modelname       & $0.60 \pm 0.04$ & $0.13 \pm 0.08$ & $0.46 \pm 0.09$ \\
 & & \modelname\ (FT) & $0.61 \pm 0.06$ & $0.15 \pm 0.08$ & $0.48 \pm 0.08$ \\
 & & Mantis (FT)      & $0.49 \pm 0.02$ & $0.01 \pm 0.01$ & $0.40 \pm 0.05$ \\
\midrule

\multirow{9}{*}{{Amigos}} &
\multirow{9}{*}{{Liking High/Low}}
  & Dummy            & $0.54 \pm 0.03$ & $0.06 \pm 0.04$ & $0.34 \pm 0.07$ \\
 & & Gen. HC           & $0.60 \pm 0.05$ & $0.28 \pm 0.08$ & $0.62 \pm 0.09$ \\
 & & EDA-spec. HC     & $0.58 \pm 0.06$ & $0.20 \pm 0.08$ & $0.56 \pm 0.09$ \\
 & & Mantis           & $0.61 \pm 0.04$ & $0.18 \pm 0.08$ & $0.51 \pm 0.07$ \\
 & & MOMENT           & $0.57 \pm 0.03$ & $0.15 \pm 0.06$ & $0.35 \pm 0.08$ \\
 & & Chronos          & $0.53 \pm 0.04$ & $0.08 \pm 0.06$ & $0.40 \pm 0.07$ \\
 & & \modelname       & $0.63 \pm 0.05$ & $0.26 \pm 0.08$ & $0.59 \pm 0.08$ \\
 & & \modelname\ (FT) & $0.66 \pm 0.12$ & $0.26 \pm 0.09$ & $0.55 \pm 0.09$ \\
 & & Mantis (FT)      & $0.58 \pm 0.04$ & $0.14 \pm 0.08$ & $0.47 \pm 0.08$ \\
\end{longtable}
\endgroup

\begin{table}[H]
\centering
\caption{Results of binary classification experiments for Time-Aware (TA) cross-validation. Results shown as \texttt{metric $\pm$ standard error}. A result reported as OoM (out of memory) means that the required memory size for the embeddings computation exceeded the computational resources available. \emph{FTC} stands for \emph{failed to converge}.}
\label{tab:binary_results_tacv}
\tiny
\setlength{\tabcolsep}{3pt}
\renewcommand{\arraystretch}{0.9}
\begin{tabular}{ll l ccc}
\toprule
\multirow{2}{*}{\textbf{Dataset}} &
\multirow{2}{*}{\textbf{Binary Task}} &
\multirow{2}{*}{\textbf{Model}} &
\multicolumn{3}{c}{\textbf{Metrics}} \\
\cmidrule(lr){4-6}
 & & & \textbf{Balanced Acc.} & \textbf{MCC} & \textbf{F1} \\
\midrule

\multirow{9}{*}{{DREAMT}} &
\multirow{9}{*}{{Deep Sleep/REM}}
  & Dummy         & $0.48 \pm 0.02$ & $0.01 \pm 0.01$ & $0.20 \pm 0.06$ \\
 & & Gen. HC        & $0.55 \pm 0.05$ & $0.09 \pm 0.10$ & $0.28 \pm 0.07$ \\
 & & EDA-spec. HC  & $0.59 \pm 0.04$ & $0.16 \pm 0.06$ & $0.37 \pm 0.07$ \\
 & & Mantis        & $0.64 \pm 0.03$ & $0.22 \pm 0.06$ & $0.40 \pm 0.06$ \\
 & & MOMENT        & $0.65 \pm 0.01$ & $0.24 \pm 0.04$ & $0.42 \pm 0.05$ \\
 & & Chronos       & $0.63 \pm 0.02$ & $0.23 \pm 0.04$ & $0.41 \pm 0.05$ \\
 & & \modelname    & $0.63 \pm 0.05$ & $0.20 \pm 0.07$ & $0.38 \pm 0.06$ \\
 & & \modelname\ (FT) & $0.67 \pm 0.03$ & $0.26 \pm 0.05$ & $0.45 \pm 0.05$ \\
 & & Mantis (FT)      & $0.65 \pm 0.04$ & $0.24 \pm 0.04$ & $0.42 \pm 0.05$ \\
\midrule

\multirow{9}{*}{{DREAMT}} &
\multirow{9}{*}{{Sleep/Wake}}
  & Dummy         & $0.50 \pm 0.01$ & $-0.01 \pm 0.01$ & $0.58 \pm 0.04$ \\
 & & Gen. HC        & $0.65 \pm 0.02$ & $0.32 \pm 0.06$ & $0.69 \pm 0.03$ \\
 & & EDA-spec. HC  & $0.69 \pm 0.01$ & $0.38 \pm 0.01$ & $0.72 \pm 0.01$ \\
 & & Mantis        & $0.73 \pm 0.01$ & $0.46 \pm 0.02$ & $0.76 \pm 0.01$ \\
 & & MOMENT        & $0.69 \pm 0.01$ & $0.39 \pm 0.01$ & $0.72 \pm 0.01$ \\
 & & Chronos       & $0.73 \pm 0.01$ & $0.47 \pm 0.01$ & $0.76 \pm 0.01$ \\
 & & \modelname    & $0.70 \pm 0.01$ & $0.41 \pm 0.02$ & $0.74 \pm 0.01$ \\
 & & \modelname\ (FT) & $0.74 \pm 0.01$ & $0.49 \pm 0.01$ & $0.78 \pm 0.01$ \\
 & & Mantis (FT)      & $0.65 \pm 0.01$ & $0.32 \pm 0.06$ & $0.69 \pm 0.03$ \\
\midrule

\multirow{7}{*}{{HHISS}} &
\multirow{7}{*}{{Stress/calm}}
  & Dummy         & $0.50 \pm 0.00$ & $-0.00 \pm 0.00$ & $0.23 \pm 0.09$ \\
 & & Gen. HC        & $0.45 \pm 0.04$ & $-0.10 \pm 0.07$ & $0.34 \pm 0.04$ \\
 & & EDA-spec. HC  & $0.55 \pm 0.03$ & $0.09 \pm 0.05$ & $0.45 \pm 0.03$ \\
 & & Mantis        & $0.51 \pm 0.03$ & $0.02 \pm 0.06$ & $0.38 \pm 0.04$ \\
 & & MOMENT        & OoM & OoM & OoM \\
 & & Chronos       & OoM & OoM & OoM \\
 & & \modelname    & $0.49 \pm 0.03$ & $-0.02 \pm 0.06$ & $0.38 \pm 0.04$ \\
\midrule

\multirow{9}{*}{{HeartS}} &
\multirow{9}{*}{{Sleep/Wake}}
  & Dummy         & $0.49 \pm 0.00$ & $-0.01 \pm 0.01$ & $0.08 \pm 0.02$ \\
 & & Gen. HC        & $0.66 \pm 0.04$ & $0.35 \pm 0.11$ & $0.39 \pm 0.08$ \\
 & & EDA-spec. HC  & $0.72 \pm 0.07$ & $0.36 \pm 0.13$ & $0.41 \pm 0.10$ \\
 & & Mantis        & $0.75 \pm 0.09$ & $0.40 \pm 0.14$ & $0.44 \pm 0.11$ \\
 & & MOMENT        & $0.69 \pm 0.06$ & $0.24 \pm 0.08$ & $0.31 \pm 0.06$ \\
 & & Chronos       & $0.74 \pm 0.06$ & $0.35 \pm 0.11$ & $0.40 \pm 0.10$ \\
 & & \modelname    & $0.73 \pm 0.09$ & $0.32 \pm 0.13$ & $0.38 \pm 0.09$ \\
 & & \modelname\ (FT) & $0.71 \pm 0.06$ & $0.36 \pm 0.12$ & $0.40 \pm 0.07$ \\
 & & Mantis (FT)      & $0.72 \pm 0.05$ & $0.36 \pm 0.13$ & $0.41 \pm 0.10$ \\
\midrule

\multirow{7}{*}{{WESAD}} &
\multirow{7}{*}{{Low/High Arousal}}
  & Dummy         & $0.55 \pm 0.05$ & $0.08 \pm 0.07$ & $0.15 \pm 0.07$ \\
 & & Gen. HC        & $0.56 \pm 0.08$ & $0.08 \pm 0.11$ & $0.26 \pm 0.06$ \\
 & & EDA-spec. HC  & $0.59 \pm 0.10$ & $0.08 \pm 0.16$ & $0.32 \pm 0.07$ \\
 & & Mantis        & $0.58 \pm 0.06$ & $0.13 \pm 0.09$ & $0.32 \pm 0.07$ \\
 & & MOMENT        & OoM & OoM & OoM \\
 & & Chronos       & $0.56 \pm 0.09$ & $0.07 \pm 0.14$ & $0.28 \pm 0.10$ \\
 & & \modelname    & $0.56 \pm 0.06$ & $0.08 \pm 0.09$ & $0.28 \pm 0.07$ \\
\midrule

\multirow{9}{*}{{WESAD}} &
\multirow{9}{*}{{Low/High Valence}}
  & Dummy         & $0.51 \pm 0.03$ & $0.01 \pm 0.04$ & $0.66 \pm 0.06$ \\
 & & Gen. HC        & $0.55 \pm 0.10$ & $0.08 \pm 0.17$ & $0.76 \pm 0.06$ \\
 & & EDA-spec. HC  & $0.54 \pm 0.10$ & $0.04 \pm 0.15$ & $0.62 \pm 0.11$ \\
 & & Mantis        & $0.61 \pm 0.04$ & $0.18 \pm 0.07$ & $0.71 \pm 0.07$ \\
 & & MOMENT        & OoM & OoM & OoM \\
 & & Chronos       & $0.45 \pm 0.06$ & $-0.10 \pm 0.10$ & $0.59 \pm 0.11$ \\
 & & \modelname    & $0.63 \pm 0.06$ & $0.18 \pm 0.09$ & $0.73 \pm 0.07$ \\
 & & \modelname\ (FT) & $0.69 \pm 0.04$ & $0.25 \pm 0.10$ & $0.77 \pm 0.08$ \\
 & & Mantis (FT)      & $0.61 \pm 0.03$ & $0.18 \pm 0.07$ & $0.71 \pm 0.07$ \\
\midrule

\multirow{9}{*}{{APSync}} &
\multirow{9}{*}{{Low/High engagement}}
  & Dummy         & $0.45 \pm 0.20$ & $0.13 \pm 0.13$ & $0.13 \pm 0.13$ \\
 & & Gen. HC        & $0.80 \pm 0.10$ & $0.32 \pm 0.16$ & $0.72 \pm 0.15$ \\
 & & EDA-spec. HC  & $0.80 \pm 0.10$ & $0.32 \pm 0.16$ & $0.72 \pm 0.15$ \\
 & & Mantis        & $0.86 \pm 0.07$ & $0.41 \pm 0.22$ & $0.82 \pm 0.10$ \\
 & & MOMENT        & $0.48 \pm 0.16$ & $-0.05 \pm 0.05$ & $0.33 \pm 0.15$ \\
 & & Chronos       & $0.61 \pm 0.20$ & $0.18 \pm 0.43$ & $0.70 \pm 0.16$ \\
 & & \modelname    & $0.89 \pm 0.11$ & $0.49 \pm 0.29$ & $0.83 \pm 0.17$ \\
 & & \modelname\ (FT) & FTC & FTC & FTC \\
 & & Mantis (FT)      & FTC & FTC & FTC \\
\midrule

\multirow{9}{*}{{EDABE}} &
\multirow{9}{*}{{Artifact}}
  & Dummy            & $0.49 \pm 0.01$ & $-0.01 \pm 0.02$ & $0.59 \pm 0.05$ \\
 & & Gen. HC           & $0.71 \pm 0.03$ & $0.44 \pm 0.06$ & $0.68 \pm 0.03$ \\
 & & EDA-spec. HC     & $0.72 \pm 0.02$ & $0.44 \pm 0.05$ & $0.70 \pm 0.04$ \\
 & & Mantis           & $0.77 \pm 0.05$ & $0.55 \pm 0.04$ & $0.79 \pm 0.03$ \\
 & & MOMENT           & $0.73 \pm 0.02$ & $0.46 \pm 0.05$ & $0.73 \pm 0.03$ \\
 & & Chronos          & $0.77 \pm 0.05$ & $0.53 \pm 0.06$ & $0.78 \pm 0.03$ \\
 & & \modelname       & $0.74 \pm 0.03$ & $0.48 \pm 0.05$ & $0.75 \pm 0.02$ \\
 & & \modelname\ (FT) & $0.81 \pm 0.01$ & $0.62 \pm 0.02$ & $0.82 \pm 0.01$ \\
 & & Mantis (FT)      & $0.76 \pm 0.02$ & $0.54 \pm 0.04$ & $0.79 \pm 0.03$ \\
\bottomrule

\end{tabular}%
\end{table}

\begin{table}[H]
\centering
\small
\caption{Model ranking under the Friedman test. Ranks are computed per
dataset--task block on balanced accuracy ($1=$ best) with mid-ranks for ties,
over the $N=14$ blocks with complete results (APSync is excluded, as
fine-tuning failed to converge). Brackets give $95\%$ percentile bootstrap
confidence intervals for the mean rank, obtained by resampling blocks with
replacement ($B=2\times10^{6}$); they are asymmetric because ranks are bounded
in $[1,k]$. These intervals are marginal, so overlap between them must not be
read as a pairwise test: significance is assessed by the Nemenyi post-hoc test
($\mathrm{CD}_{\alpha=0.05}=3.21$), and $^{\dagger}$ marks methods ranked
significantly below the top-ranked method. The omnibus test rejects the null
of equal performance ($\chi^{2}_{F}(8)=57.20$, $p=1.6\times10^{-9}$;
Iman--Davenport $F(8,104)=13.57$, $p=2.7\times10^{-13}$).}
\label{tab:friedman-ranks}
\begin{tabular}{clcc}
\toprule
\# & Method & Avg.\ rank [$95\%$ CI] & Mean bal.\ acc. \\
\midrule
1 & \modelname (fine-tuned)      & $\mathbf{2.11}$ $[1.36, 3.00]$   & $.707$ \\
2 & Mantis                & $3.32$ $[2.39, 4.46]$            & $.681$ \\
3 & \modelname       & $3.93$ $[3.29, 4.61]$            & $.678$ \\
4 & Chronos               & $4.36$ $[3.00, 5.82]$            & $.649$ \\
5 & Mantis (fine-tuned)   & $4.54$ $[3.61, 5.43]$            & $.667$ \\
6 & HC EDA-spec.          & $5.43$ $[4.36, 6.36]^{\dagger}$  & $.657$ \\
7 & MOMENT                & $5.75$ $[4.68, 6.71]^{\dagger}$  & $.649$ \\
8 & HC Gen.               & $7.11$ $[6.39, 7.75]^{\dagger}$  & $.625$ \\
9 & Dummy                 & $8.46$ $[7.75, 8.93]^{\dagger}$  & $.504$ \\
\bottomrule
\end{tabular}
\end{table}

\end{document}